\documentclass[10pt,journal,compsoc]{IEEEtran}

\ifCLASSOPTIONcompsoc
  \usepackage[nocompress]{cite}
  \usepackage{amsmath,amsfonts}
\usepackage{algorithmic}
\usepackage[ruled,linesnumbered]{algorithm2e}
\usepackage{array}
\usepackage[caption=false,font=normalsize,labelfont=sf,textfont=sf]{subfig}
\usepackage{textcomp}
\usepackage{stfloats}
\usepackage{url}
\usepackage{verbatim}
\usepackage{graphicx}
\usepackage{booktabs}
\usepackage{enumitem}
\usepackage[table,xcdraw]{xcolor}
\else
    \usepackage{cite}
\fi

\usepackage[sort]{natbib} 
\setcitestyle{numbers,square}

\ifCLASSINFOpdf
\else
\fi
\usepackage{xcolor}
\usepackage{multirow}
\begin{document}
\definecolor{bxy}{rgb}{1,0.5,0}
\definecolor{zpf}{rgb}{1,0,0}
\newcommand{\gpt}[1]{\textcolor{orange}{\textbf{GPT:} #1}}
\newcommand{\zpf}[1]{\textcolor{red}{\textbf{ZPF:} #1}}
\newcommand{\qms}[1]{\textcolor{blue}{#1}}
\newcommand{\bxy}[1]{\textcolor{orange}{#1}}

%
\title{Towards Robust Unsupervised Attention Prediction in Autonomous Driving}
%
%
%
%

\author{
        Mengshi Qi,~\IEEEmembership{Member,~IEEE},
        Xiaoyang Bi,
        Pengfei Zhu,
        Huadong Ma,~\IEEEmembership{Fellow,~IEEE}
\thanks{This work is partly supported by the Funds for the NSFC Project under Grant 62202063, Beijing Natural Science Foundation (L243027). (\emph{Corresponding author: Mengshi Qi~(email:~qms@bupt.edu.cn)})}
\thanks{M. Qi, X. Bi, P. Zhu, and H. Ma are with State Key Laboratory of Networking and Switching Technology, Beijing University of Posts and Telecommunications, China.}
}
\IEEEtitleabstractindextext{%
\begin{abstract}
Robustly predicting attention regions of interest for self-driving systems is crucial for driving safety but presents significant challenges due to the labor-intensive nature of obtaining large-scale attention labels and the domain gap between self-driving scenarios and natural scenes. These challenges are further exacerbated by complex traffic environments, including camera corruption under adverse weather, noise interferences, and central bias from long-tail distributions. To address these issues, we propose a robust unsupervised attention prediction method. An Uncertainty Mining Branch refines predictions by analyzing commonalities and differences across multiple pre-trained models on natural scenes, while a Knowledge Embedding Block bridges the domain gap by incorporating driving knowledge to adaptively enhance pseudo-labels. Additionally, we introduce RoboMixup, a novel data augmentation method that improves robustness against corruption through soft attention and dynamic augmentation, and mitigates central bias by integrating random cropping into Mixup as a regularizer. To systematically evaluate robustness in self-driving attention prediction, we introduce the \emph{DriverAttention-C} benchmark, comprising over 100k frames across three subsets: BDD-A-C, DR(eye)VE-C, and DADA-2000-C. Our method achieves performance equivalent to or surpassing fully supervised state-of-the-art approaches on three public datasets and the proposed robustness benchmark, reducing relative corruption degradation by 58.8\% and 52.8\%, and improving central bias robustness by 12.4\% and 11.4\% in KLD and CC metrics, respectively. Code and data are available at \text{https://github.com/zaplm/DriverAttention}.
\end{abstract}


}

\maketitle

\IEEEdisplaynontitleabstractindextext

%
\IEEEpeerreviewmaketitle

\IEEEraisesectionheading{\section{Introduction}\label{sec:introduction}}
\IEEEPARstart{I}{n} recent years, significant advancements in autonomous driving have heightened interest in predicting attention regions for self-driving systems~\cite{baee2021medirl,pal2020looking} within the research and industry community. Predicted attention regions offer vital contextual information, aiding autonomous driving systems in identifying key areas within traffic scenes~\cite{xu2020explainable,qi2020stc,qi2019ke}. Crucially, these key areas often encompass the highest risk zones, where minor perception errors can significantly jeopardize driver safety~\cite{kendall2017uncertainties}. Consequently, successful prediction of attention areas allows for the reallocation of computational resources. This reallocation enhances perception accuracy in critical zones, thereby mitigating driving risks and bolstering the explainability and reliability of autonomous driving systems~\cite{palazzi2018predicting}.

\begin{figure}
    \centering
    \includegraphics[width=0.95\linewidth]{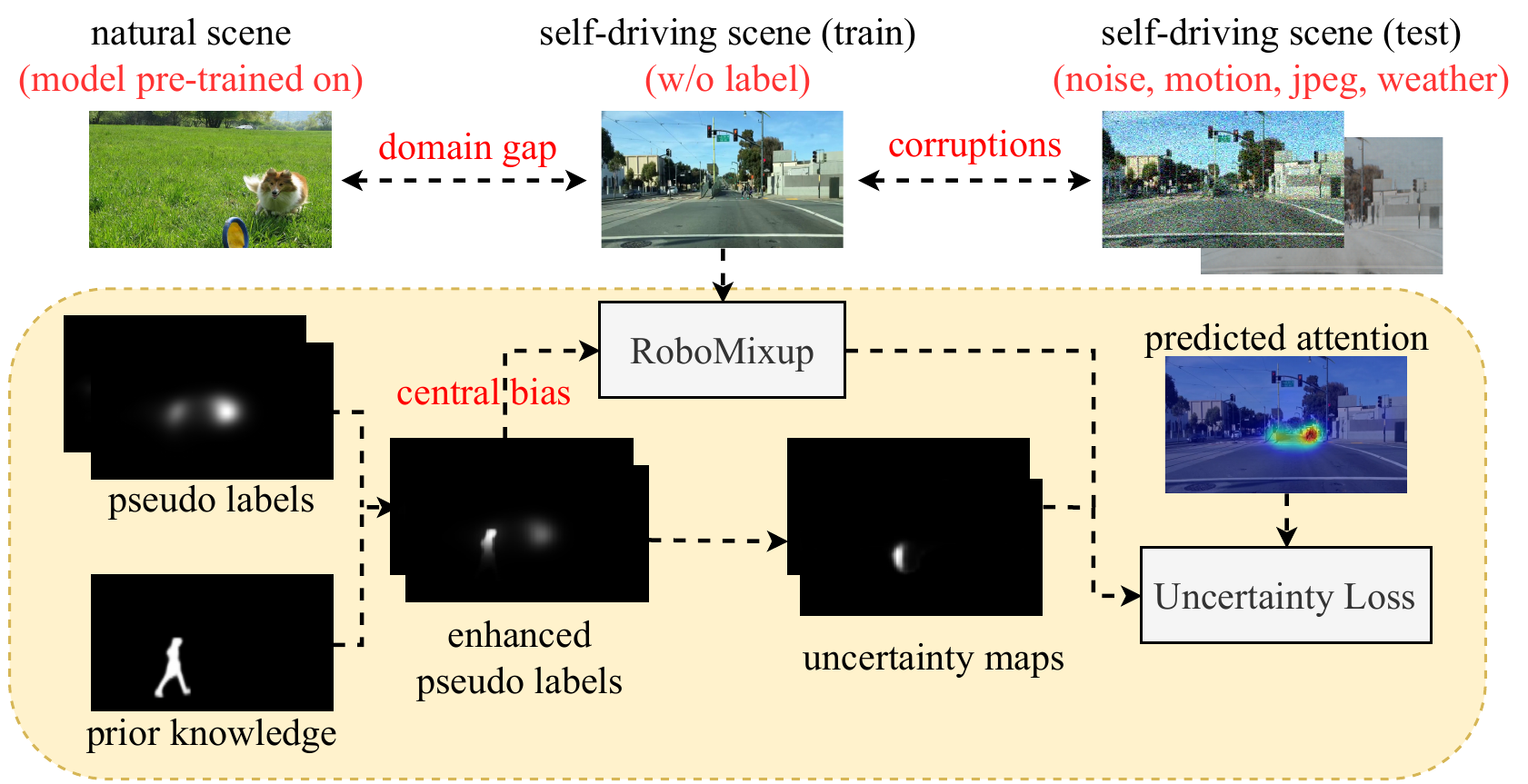}
    \vspace{-3mm}
    \caption{Illustration of the proposed unsupervised attention prediction in self-driving. Our method bypasses the need for ground truth labels from traffic datasets, by leveraging pseudo-labels generated from models pre-trained on natural scenes. These pseudo-labels are refined through the knowledge embedding and, together with the images, are processed by the RoboMixup to address corruption and central bias. Finally, the uncertainty map and loss guide the model in learning attention regions.}
    \label{fig:motivation}
    \vspace{-5mm}
\end{figure}

A variety of datasets~\cite{xia2018predicting,alletto2016dr,fang2021dada} and methodologies~\cite{palazzi2018predicting,xia2018predicting,baee2021medirl,kim2020advisable} have been introduced to tackle the task of predicting attention in self-driving. Despite their promising performance, these methods rely on fully-supervised training using large-scale labeled datasets, which are difficult and unreliable to construct. For instance, the DR(eye)VE~\cite{alletto2016dr} dataset, widely used in self-driving research, was compiled over two months by recording eight drivers alternately navigating the same route to gather fixation data. However, averaging the attention data from eight drivers into a single video can result in inaccurate attention targeting. Another significant challenge is the substantial disparity between collected data and real-world environments. The BDD-A~\cite{xia2018predicting} dataset, another dataset for self-driving, was created by having 45 participants watch a recorded video and envision themselves as the drivers. However, this simulation approach inevitably introduces inconsistencies with real-world conditions for human labeling. Consequently, current fully-supervised methods are prone to biases in public datasets, making them difficult to adapt to new environments. Moreover, large-scale pre-trained models have shown strong capabilities in representation learning, offering benefits for numerous downstream tasks. However, bridging the domain gap between specific situations (\textit{e.g.}, self-driving scenes) and the domains of pre-trained models (\textit{e.g.}, natural scenes) remains a challenge.

Furthermore, autonomous driving based on RGB cameras faces challenges in complex traffic environments, where maintaining high awareness of the scene is crucial for safety~\cite{almalioglu2022deep}. The complexity of traffic scenes presents challenges to robustness: \textbf{1)} camera inputs corrupted by adverse conditions, such as noise interferences, blur, extreme weather, and digital distortions~\cite{hendrycks2019benchmarking, wang2021human}; and \textbf{2)} the current self-driving attention prediction datasets exhibit long-tail distributions~\cite{xia2018predicting, palazzi2018predicting}, leading models to predict attention predominantly at the center of the roadway, overlooking critical scene objects, a phenomenon known as central bias~\cite{xia2018predicting, palazzi2018predicting}. However, the above-mentioned issues pose a serious threat to driving safety. For example, motion blur can occur when a vehicle moves swiftly, and the camera fails to compensate adequately. 



To tackle the issues outlined above, we introduce an innovative robust unsupervised framework for predicting attention in self-driving, meaning \textbf{1)} the exclusion of ground-truth labels from self-driving datasets, \textbf{2)} the exclusive use of pseudo-labels from models pre-trained on natural scene datasets, \textbf{3)} training a model within the source domain and adapting it for the target domain samples (specifically, from natural to traffic scenes) in accordance with unsupervised domain adaptation principles~\cite{kang2019contrastive}. To be specific, our model is distinguished by two innovative components: An Uncertainty Mining Branch to leverage uncertainties in pseudo-labels by aligning diverse distributions, thereby enhancing result reliability; A Knowledge Embedding Block designed to infuse traffic knowledge into the natural domain, segmenting key traffic objects using Mask-RCNN~\cite{he2017mask} pre-trained on MS-COCO~\cite{lin2014microsoft} to improve the attention region of each pseudo-label. Furthermore, to systematically analyze and improve the robustness of self-driving attention prediction tasks, \textbf{1)} we constructed a new corruption dataset named \emph{DriverAttention-C}, which stems from three existing widely-adopted datasets: BDD-A-C, DR(eye)VE-C, and DADA-2000-C~\cite{xia2018predicting, alletto2016dr, fang2021dada}, and includes 115,332 frames within four categories of corruptions: noise, blur, digital, and weather. Specifically, we utilized Cycle-GAN~\cite{CycleGAN2017} to generate fog and snow data to simulate realistic extreme weather conditions. \textbf{2)} We benchmarked the performance of state-of-the-art models on these corruption datasets. \textbf{3)} We propose a pioneering data augmentation strategy, RoboMixup, which combines soft attention-based mixup with a dynamic augmentation strategy against the corruption. Additionally, it improves robustness against central bias by mitigating the central distribution of pseudo-labels through the incorporation of random cropping into Mixup as a regularizer.

More importantly, it should be mentioned that this paper is an extension of our ICCV conference paper~\cite{zhu2023unsupervised}.~Compared to the original version, we present a new knowledge mining strategy to automate the extraction of prior knowledge in traffic scenarios without manual specification, thereby enhancing generalization. Moreover, we introduce the RoboMixup as a new data augmentation to address the corruption and central bias in self-driving attention prediction. Additionally, we generate a new corruption dataset, \emph{DriverAttention-C}, and conduct extensive experiments and detailed analysis. Last but not least, we propose an explainable autonomous driving decision-making method based on attention prediction maps to demonstrate the importance of the attention prediction in the real application. Numerous qualitative and quantitative experiments provide evidence of the superior robustness of our methods, laying a foundation for enhancing the safety of autonomous driving applications.

In summary, our contributions can be listed as follows:

\par\textbf{(1)} We propose an innovative robust unsupervised framework for predicting attention regions in self-driving that does not depend on any traffic dataset labels. 

\par\textbf{(2)} We introduce a new uncertainty mining branch, which generates credible attention maps by assessing similarities and differences among readily available pseudo-labels from models pre-trained on natural scenes.

\par\textbf{(3)} We design a novel knowledge embedding block, which refines pseudo-labels by automatically integrating the traffic knowledge, effectively bridging the domain gap between autonomous driving and various common domains.

\par\textbf{(4)} We present a new data augmentation method, RoboMixup that significantly improves the corruption robustness and mitigate the central bias.

\par\textbf{(5)} We generate the first corruption dataset in self-driving attention prediction, \emph{DriverAttention-C}, and comprehensive testing on four public benchmarks shows our method achieves comparable or superior results to fully-supervised state-of-the-art approaches, highlighting its effectiveness and the superior robustness.

\section{Related Work}


\noindent{\bf Self-Driving Attention Prediction.}~The advent of deep learning has spurred numerous initiatives into self-driving attention prediction~\cite{xia2018predicting,palazzi2018predicting,fang2021dada, chen2023fblnet}. Palazzi \textit{et al.}~\cite{palazzi2018predicting} used a multi-branch video analysis method for prediction of driver attention. Baee \textit{et al.}~\cite{baee2021medirl} enhanced attention prediction accuracy through an inverse reinforcement learning approach. However, this prior research mostly depended on extensively annotated datasets~\cite{fang2021dada,xia2018predicting,alletto2016dr} collected either in-lab or in-vehicle. For example, the DR(eye)VE~\cite{alletto2016dr} dataset, an in-vehicle collection, features multiple segments documenting changes in driver attention. BDD-A~\cite{xia2018predicting} and DADA-2000~\cite{fang2021dada}, as in-lab datasets, compile synthesized attention shifts from volunteers across over 1000 clips. Addressing the unreliability of self-driving datasets, our model pioneers unsupervised attention prediction in self-driving by using pseudo-labels from models pre-trained on natural scenes.

\noindent{\bf Saliency Detection.}~Predicting saliency regions in images or videos~\cite{min2019tased,droste2020unified} can approximate human’s visual attention. It has been used to evaluate the explainability of deep models~\cite{xu2020explainable} and to assist other tasks, \emph{i.e.}, photo cropping~\cite{wang2017deep}, scene understanding~\cite{qi2019attentive,qi2021semantics,qi2018stagnet} and object segmentation~\cite{xu2020explainable}. The early effort used a fusion strategy~\cite{fan2020bbs}, merging semantic information from various levels without addressing semantic discrepancies. To address this challenge, Xie \textit{et al.} introduced progressive feature aggregation techniques based on the feature pyramid network (FPN)~\cite{xie2022pyramid}. Afterwards, Li \textit{et al.} developed a dynamic search process to enable adaptive feature selection at the pixel level~\cite{li2023robust}. However, most existing datasets~\cite{jiang2015salicon,wang2018revisiting} and methods~\cite{min2019tased,cornia2016deep,cornia2018predicting,droste2020unified, xie2022pyramid, li2023robust} are mainly focusing on natural scenes or common objects, not specially tailored into self-driving scenarios. In this work, we propose an uncertainty mining branch, which utilizes multi-scale features with progressive aggregation. Meanwhile, a knowledge embedding strategy is introduced to bridge the domain gap between natural scenes and self-driving situations.


\begin{figure*}
    \centering
    \includegraphics[width=0.97\linewidth]{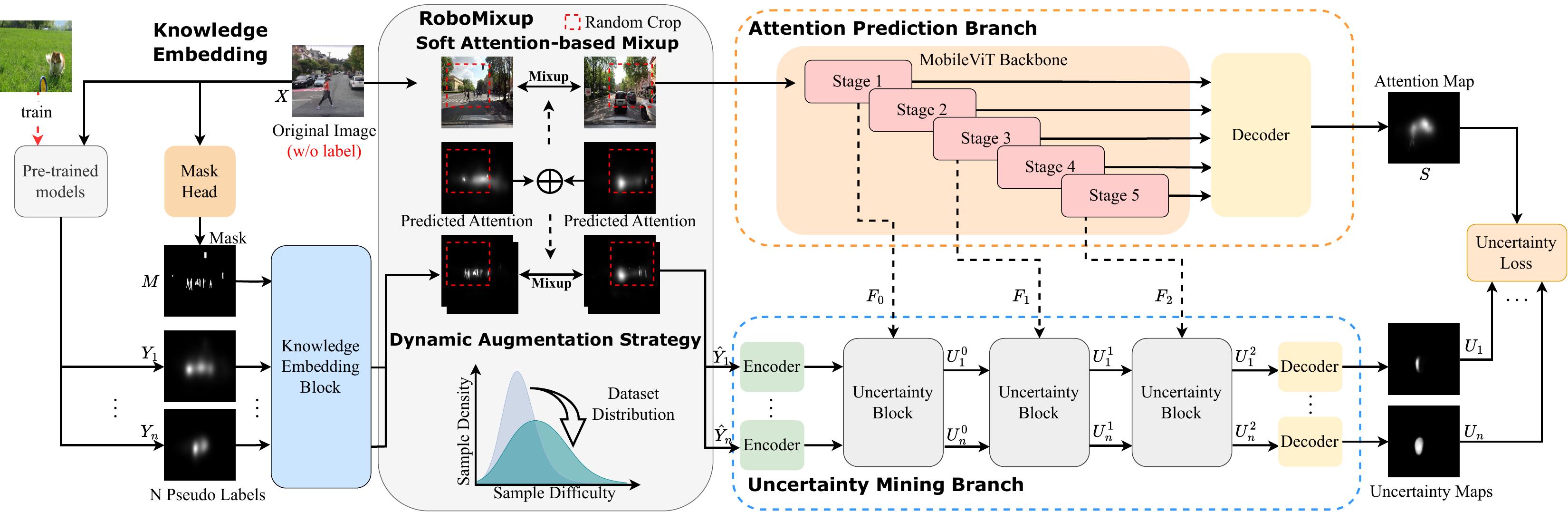}
    \vspace{-3mm}
    \caption{Overview of our proposed model. We utilize pseudo-labels from models pre-trained on natural scene datasets for unsupervised training in our approach. The Knowledge Embedding Block (KEB) is designed to integrate additional semantic information into the self-driving scenario. Our proposed data augmentation method, RoboMixup, strengthens the model's robustness against corruption and central bias by combining soft attention or random cropping with a dynamic augmentation strategy, generating realistic and challenging samples. The Attention Prediction Block (APB), using the Mobile-ViT~\cite{mehta2022mobilevit} backbone, consists of five stages of image feature extraction, each feeding its output to the decoder. Features from stages 1, 2, and 4 are directed to three Uncertainty blocks for multi-scale feature fusion. The Uncertainty Mining Block (UMB) uses multi-scale feature fusion and mining across multiple pseudo-labels to create an uncertainty map for each and then optimize with the uncertainty loss.} 
    \label{fig:overview}
    \vspace{-2mm}
\end{figure*}

\noindent{\bf Uncertainty Estimation.}~Early works in deep learning mainly focus on model uncertainty via Bayesian inference~\cite{daxberger2021laplace,gal2016dropout,kendall2017uncertainties}, which assumes that model parameters follow a specific prior distribution and then infers the posterior distribution of weights. Common approaches include variational Bayesian inference, Markov chain Monte Carlo, and Laplace approximation~\cite{daxberger2021laplace}, etc. A pioneer work is that Gal and Ghahramani~\cite{gal2016dropout} adopted dropout to represent model uncertainty. Besides, data uncertainty is typically estimated by an auxiliary model, which shares inputs, intermediate features, and labels with the main network~\cite{kendall2017uncertainties,kendall2018multi}. Kendall \textit{et al.} utilized data uncertainty as weights in multi-task learning~\cite{kendall2018multi} to balance multiple loss functions. In the field of self-driving attention prediction, we are the first to introduce an uncertainty mining branch to estimate the commonality and distinction between multiple pseudo-labels, and then produce plausible attention maps.



\noindent\textbf{Robustness.}~The robustness is critically important for ensuring the safety and reliability of autonomous driving systems~\cite{almalioglu2022deep}. However, common corruptions and disruptions pose significant challenges across various tasks~\cite{hendrycks2019benchmarking}. Hendrycks and Dietterich categorize common corruptions into four types, \emph{noise, blur, weather changes, and digital distortions}, and collect the corruption benchmark in object classification~\cite{hendrycks2019benchmarking}. Subsequent works have extended corruption benchmarking to video classification, embodied navigation, semantic segmentation, and pose estimation~\cite{wang2021human}. In this work, we are the first to construct a corruption dataset named \emph{DriverAttention-C} and benchmark the robustness of attention prediction against common corruptions caused by camera anomalies and adverse weather conditions. Additionally, we propose integrating soft attention-based mixup with a dynamic augmentation strategy to enhance the robustness. Besides, current attention prediction datasets exhibiting long-tail distributions~\cite{xia2018predicting, palazzi2018predicting}, result in central bias. This bias causes models to concentrate their attention on the roadway’s center, often overlooking critical scene elements~\cite{xia2018predicting, palazzi2018predicting}. Cornia~\textit{et al.} focus on utilizing a learned prior to incorporate center bias~\cite{cornia2016deep, cornia2018predicting}. Xia~\textit{et al.} advocated for weighted sampling based on the central bias distribution in the ground truth~\cite{xia2018predicting}. In this work, we incorporate random cropping into mixup as a regularizer, which significantly improves robustness against the central bias issue.


\noindent\textbf{Data Augmentation.}~Traditional data augmentation strategies typically apply only to the image domain, such as cropping, rotations, and random flipping~\cite{chen2022transmix}. Recently, Mixup and its variants as a data augmentation technique have been widely proven to effectively enhance the robustness of the model. Zhang \textit{et al.}~\cite{zhang2018mixup} are the first to propose Mixup, a method that employs linear interpolation on both input and label spaces. Subsequent improvements have refined the sample blending strategy, such as introducing learnable parameters, interpolating in feature space and local regions~\cite{chen2022transmix, pinto2022regmixup}. Insipred by Mixup, we propose a new data augmentation strategy, RoboMixup, which goes beyond sample interpolation by introducing a dynamic augmentation strategy and utilizing random cropping as a novel regularization technique to strengthen the robustness.


\section{Method}
\subsection{Overview}

Figure~\ref{fig:overview} provides an overview of the robust unsupervised driving attention prediction network we proposed. The architecture is composed of several key components, including the Attention Prediction Branch (APB), the Knowledge Embedding Block (KEB), the Uncertainty Mining Branch (UMB), and the RoboMixup method.

Our method predicts self-driving attention through an unsupervised learning paradigm. While pseudo-labels generated by a single source model pre-trained on natural scene datasets can be used for training, the domain gap between natural environments and self-driving scenarios introduces significant uncertainty. Single-source pseudo-labels often exhibit distinct distributions, with certain regions contributing to elevated uncertainty. Inspired by recent advancements in uncertainty estimation~\cite{kendall2017uncertainties,kendall2018multi}, we improved prediction accuracy and robustness by modeling uncertainty using pseudo-labels from multiple sources. To address the lack of autonomous driving knowledge in pseudo-labels transferred from natural domains, we incorporated a Knowledge Embedding Block (KEB) to refine input pseudo-labels, enhancing final predictions. We also introduced the Uncertainty Mining Branch (UMB), which utilizes multiple Uncertainty Blocks (UB) to iteratively analyze similarities and variations among noisy labels, producing pixel-level uncertainty maps. Additionally, we proposed RoboMixup, a novel data augmentation technique that improves corruption robustness and mitigates central bias.


\textbf{Problem Formulation.}~For an input RGB frame $X\in \mathbb{R}^{H\times W\times 3}$, APB adopts a pyramid feature extraction approach inspired by PSPNet~\cite{zhao2017pyramid}, producing features across five hierarchical levels. Features $F$ from the 1st, 2nd, and 4th stages, denoted as $\{F^0, F^1, F^2\}$, are passed to the Uncertainty Mining Branch (UMB) to analyze the uncertainty of pseudo-labels. The APB is structured following U-Net~\cite{williams2023unified}, where the features from the final layer are fed into a decoder. These features are then concatenated with features of corresponding resolutions, and the resulting attention prediction map is generated as $S\in \mathbb{R}^{H\times W\times 1}$. 
Additionally, a knowledge enhancement process is applied to adapt pseudo-labels for autonomous driving scenarios using a pre-trained Mask Head. The UMB then receives $N$ knowledge-enhanced pseudo-labels, $\hat{Y}=\{\hat{Y}_1, \cdots, \hat{Y}_N\}$, as input and produces corresponding uncertainty maps, each having the same dimensions as the predicted attention map $S$. These pseudo-labels are further fused with three levels of APB features to generate the uncertainty maps $U=\{U_1, \cdots, U_N\}$. 
To improve the training process, the RoboMixup technique is employed, which generates augmented data pairs $(\tilde{I}, \tilde{Y})$ by combining image-pseudo labels pairs $(I, \hat{Y})$ and the corresponding predicted attention map $S$. The model is subsequently optimized on these augmented data pairs using an uncertainty-based loss function.

\subsection{Knowledge Embedding Block (KEB)}
\label{sec:kes}

Humans naturally use prior knowledge to interpret and identify relevant objects in visually complex scenes~\cite{cornia2016deep}. Inspired by this capability, we designed the Knowledge Embedding Block (KEB) to incorporate prior traffic knowledge and reduce the domain gap between natural scenes and self-driving environments. Using an off-the-shelf Mask R-CNN model pre-trained on the MS-COCO dataset~\cite{lin2014microsoft}, we generate a binary segmentation map \( M \) by merging masks across relevant categories. To keep the method unsupervised, we freeze the parameters of Mask R-CNN with open-source checkpoints. Representative traffic-related objects, such as pedestrians, signals, bicycles, motorcycles, and traffic signs (\textit{e.g.,} stop signs and road signs), are identified as prior knowledge through a knowledge mining strategy. This traffic-specific prior knowledge is then incorporated into the pseudo-label generation process via a knowledge embedding mechanism, enhancing the model's adaptability to autonomous driving scenarios.

\begin{figure}
    \centering
    \includegraphics[width=0.95\linewidth]{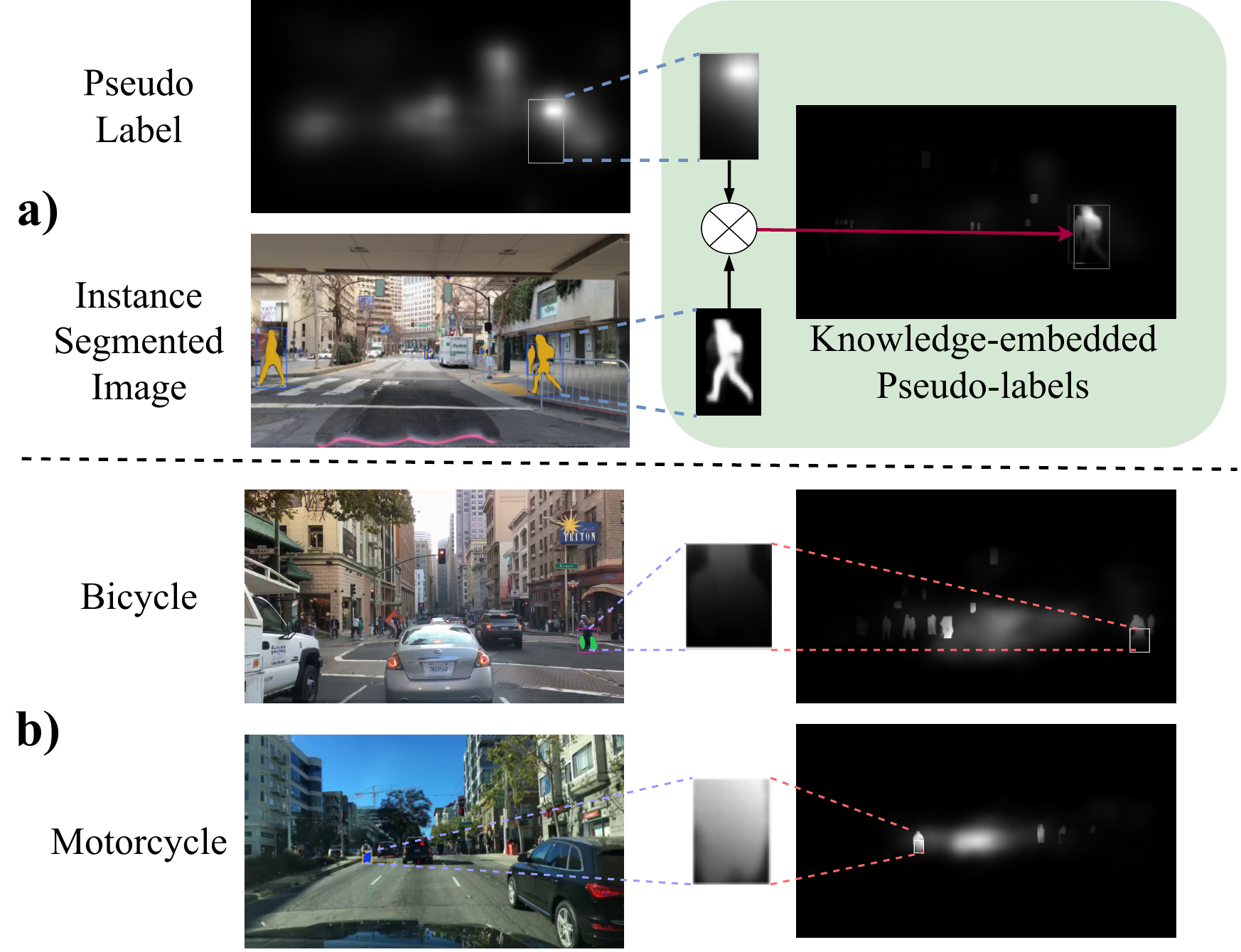}
    \vspace{-3mm}
    \caption{Illustration of the knowledge embedding strategy: a) the process of knowledge embedding for a single pseudo-label, where the salient region can be enhanced by adding the self-driving-related instance (\textit{e.g.} pedestrian) where the operator $\otimes$ means the operation in Eq.~(\ref{eqo:prior}); b) two other examples of knowledge embedding for bicycles and motorcycles.}
    \label{fig:kes}
    \vspace{-2mm}
\end{figure}

\textbf{Knowledge Mining Strategy.}
\label{sec:kms}
For all segmented categories, we first count the occurrences of each category in the dataset and sort them in descending order to obtain \( C = \{c_1, c_2, \ldots, c_n\} \). We focus on the most frequent categories, as the remaining categories are likely to be misclassified due to their small proportion in the dataset and the fact that the Mask R-CNN model we used is not trained on traffic scenes. Specifically, we choose the following categories:
\begin{equation}
\hat{C} = \left\{ c_i \mid i \leq \min \left\{k \mid T(k) \geq p\% \times T\left(\left|\bar{C}\right|\right) \right\} \right\},
\end{equation}
where \( p \) is the coverage threshold, and \( T(k) \) represents the sum of the instance numbers of the first \( k \) categories in C. Pretrained attention prediction models for natural scenes lack the recognition capabilities for traffic-specific objects (e.g., stop signs), leading to an attention distribution bias across various objects (\emph{i.e.,} draw more attention on daily objects than traffic items). We leverage this bias for knowledge mining by calculating the mean attention for all pseudo-label instances as:
\begin{equation}
V_{c_i} = \frac{1}{n_{c_i}} \sum_{j=1}^{n_{c_i}} \left( \frac{\mu_P \cdot M_{ij}}{M_{ij}} \right),
\end{equation}
where \( \mu_P \) denotes the mean of pseudo-labels generated by all pretrained models on natural scenes, \( M_{ij} \) is the binary segmentation mask for the \( j^{\text{th}} \) instance of the category \( c_i \), and $n_{c_i}$ represents the instance count for the category ${c_i}$. As a result, we mine the following categories as prior knowledge:
\begin{equation}
\tilde{C} = \left\{ c_i \mid V_{c_i} < \eta \sum_{j=1}^{k} V_{c_j} \right\},
\end{equation}
where \( \eta \) is the proportion factor. The process of generating the segmentation map with prior knowledge can be formally expressed as:
\begin{equation}
\hat{M} = M \cdot \mathbf{1}_{\tilde{C}},
\end{equation}
where \( \mathbf{1}_{\tilde{C}} \) is a binary mask that indicates the prior categories, with a value of 1 for categories in \( \tilde{C} \) and 0 for categories not in \( \tilde{C} \)




\textbf{Knowledge Embedding Technique.}
We investigated two approaches for integrating prior knowledge from the segmentation map into pseudo-labels: (1) concatenating the segmentation map along the channel dimension of the pseudo-labels, and (2) fusing the segmentation map into a single-channel representation. In the first approach, each pseudo-label is concatenated with the binary segmentation mask along the channel axis and subsequently passed to the UMB, enabling the model to adaptively learn the relationships. In the second approach, the segmentation map is combined with each pseudo-label using the following equation:
\begin{equation}
    {{\hat{Y}}_{n}} = {{Y}_{n}} \cdot \left( {\hat{M}} + \alpha \right),
\label{eqo:prior}
\end{equation}
where \( \alpha \) serves as an adjustment parameter, \( Y_{n} \) denotes the \( n \)-th pseudo-label, and \( \hat{M} \) is the segmentation map embedding prior knowledge associated with the input image. 
Following this knowledge embedding process, the pseudo-labels are enhanced at the pixel level, allowing the model to more effectively and robustly identify key traffic-related objects in self-driving environments.

    


\subsection{Uncertainty Mining Branch (UMB)}

The Uncertainty Mining Branch (UMB) is designed to extract uncertainty from multi-source pseudo-labels generated by several pre-trained models. Notably, these models are trained on natural scene datasets rather than self-driving data. For instance, ML-Net~\cite{cornia2016deep}, SAM~\cite{cornia2018predicting}, and UNISAL~\cite{droste2020unified} are trained on SALICON~\cite{jiang2015salicon}, while TASED-Net~\cite{min2019tased} is pre-trained on DHF-1K~\cite{wang2018revisiting}. As illustrated in Figure~\ref{fig:overview}, the Uncertainty Block (UB) is introduced to facilitate information exchange between pseudo-labels and multi-scale features extracted by the APB. Each Uncertainty Block leverages non-local self-attention mechanisms and a merge/split design~\cite{wang2018non,wang2022multi} to achieve this.
Specifically, for the $n$-th knowledge-embedded pseudo-label $\hat{Y_n}\in \mathbb{R}^{H\times W\times 1}$, we first process it through a convolutional layer followed by a downsampling operation, reducing its spatial dimensions to one-fourth of the original size. The downsampled pseudo-label is then passed through a residual block~\cite{he2016deep}, enabling information exchange between pseudo-labels and feature maps derived from other sources at the same stage. The processed outputs are concatenated with the input pseudo-labels and passed through a non-local self-attention mechanism to generate a coarse uncertainty map $U_n$ for the $n$-th pseudo-label. This process is mathematically expressed as:
\begin{equation}
    {{U}_{n}^{0}}={{f}_{attn}^0}\left( \mathrm{Concat}\left( \hat{Y_{1}},\cdots ,\hat{Y}_{n},{{F}^{0}} \right) \right)+\hat{Y}_{n},
\end{equation}
\noindent where the superscripts indicate the stage index, and $f_{attn}^t(\cdot)$ represents the non-local self-attention function. The uncertainty map $U_{n}^{0}$ is iteratively refined across stages, yielding:
\begin{equation}
    {{U}_{n}^{t+1}}={{f}_{attn}^t}\left( \mathrm{Concat}\left( U_{1}^{t},\cdots ,U_{N}^{t},{{F}^{t}} \right) \right)+U_{n}^{t}.
\end{equation}
After passing through three Uncertainty Blocks, the fine-grained uncertainty map ${{U}_{n}^{2}}\in \mathbb{R}^{\frac{H}{4}\times \frac{W}{4}\times 1}$ is produced. This map is subsequently upsampled within the decoder to match the original input size, resulting in $U_{n}\in \mathbb{R}^{H\times W\times 1}$.

\subsection{Uncertainty Loss Function}

We treat the predicted attention map $S$ as a spatial distribution and normalize the generated pseudo-labels accordingly. To achieve this, a spatial softmax layer is applied after the APB. Drawing inspiration from the uncertainty loss framework in~\cite{kendall2017uncertainties}, we model each pseudo-label map $\hat{Y}_n \in \mathbb{R}^{H \times W \times 1}$ as a Boltzmann distribution under Bayesian theory. The probability of $S$ conditioned on the pseudo-label $\hat{Y}_n$ is given by:
\begin{equation}
    p(\hat{Y}_n | S, u_n) = \prod_i \mathrm{Softmax} \left(\frac{S_i}{u_n^2}\right),
\label{eqo:base}
\end{equation}
where \( u_n = \frac{1}{H \times W} \sum_{i}^{H \times W} U_n^i \) represents the uncertainty estimation for the $n$-th pseudo-label, \( i \) denotes the pixel index of $S$, and \( u_n \) serves as a temperature parameter controlling the distribution’s flatness. 
The negative log-likelihood for the pseudo-label map is calculated as:
\begin{equation}
    \begin{aligned}
         -\log p(\hat{Y}_n | S, u_n) &= -\sum_i \frac{S_i}{u_n^2} + \log \sum_i \exp \left(\frac{S_i}{u_n^2}\right) \\
        &\approx \frac{\mathcal{L}_{\mathrm{CE}}(S, \hat{Y}_n)}{u_n^2} + \log(u_n),
    \end{aligned}
    \label{eqo:origin}
\end{equation}
where \( \mathcal{L}_{\mathrm{CE}}(S, \hat{Y}_n) \) denotes the spatial cross-entropy loss. To improve numerical stability during training, we predict the log variance \( e_n = \log(u_n^2) \) as suggested in~\cite{kendall2018multi}. The uncertainty loss is reformulated as:
\begin{equation}
    \mathcal{L}_{unc}(S, u_n, \hat{Y}_n) = \mathcal{L}_{\mathrm{CE}}(S, \hat{Y}_n) \cdot \exp(-e_n) + \frac{1}{2}e_n.
    \label{eq:uncertainty}
\end{equation}
We further express the cross-entropy loss \( \mathcal{L}_{\mathrm{CE}}(S, \hat{Y}_n) \) as:
\begin{equation}
    \begin{aligned}
        \mathcal{L}_{\mathrm{CE}}(S, \hat{Y}_n) &= -\sum_i \hat{Y}_{n,i} \log(S_i) \\
        &= -\sum_i \hat{Y}_{n,i} \log(S_i) + H(\hat{Y}_n) - H(\hat{Y}_n) \\
        &= \sum_i \hat{Y}_{n,i} (\log(\hat{Y}_{n,i}) - \log(S_i)) - H(\hat{Y}_n) \\
        &= \mathcal{L}_{\mathrm{KLD}}(\hat{Y}_n, S) - H(\hat{Y}_n),
    \end{aligned}
    \label{eq:kld}
\end{equation}
where \( \mathcal{L}_{\mathrm{KLD}}(\hat{Y}_n, S) = \sum_i \hat{Y}_{n,i} \left(\log(\hat{Y}_{n,i}) - \log(S_i)\right) \) represents the KL-divergence between the pseudo-label distribution and the predicted attention map distribution, and \( H(\hat{Y}_n) \) denotes the information entropy of \( \hat{Y}_n \), which remains constant during optimization. 
Extending the calculation across all $N$ pseudo-labels, the total uncertainty loss is derived as:
\begin{equation}
    \mathcal{L}_{unc} = \sum_{n=1}^N \left\{\mathcal{L}_{\mathrm{KLD}}(\hat{Y}_n, S) \cdot \exp(-e_n) + \frac{1}{2}e_n \right\}.
\label{eq:total_loss}
\end{equation}
It is worth noting that our KL-divergence-based uncertainty loss differs from prior works~\cite{palazzi2018predicting} by assuming a spatial distribution rather than a single per-channel counterpart. This assumption is critical for deriving Eq.~(\ref{eq:kld}) and ensures alignment with the spatial nature of our task.

\begin{figure*}[ht]
  \centering
  \captionsetup[subfloat]{font=scriptsize, justification=centering}
  \subfloat[Soft Attention-based Mixup\label{fig:mixup_visual}]{\includegraphics[width=0.5\textwidth]{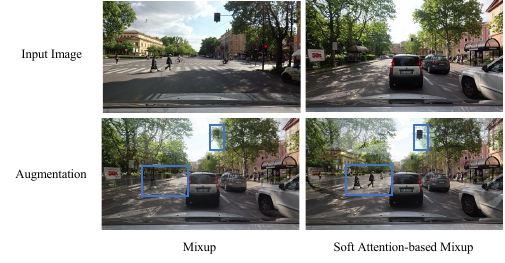}}
    \hspace{2em} 
  \subfloat[Dynamic Augmentation\label{fig:distribution_comp}]{\includegraphics[width=0.42\textwidth]{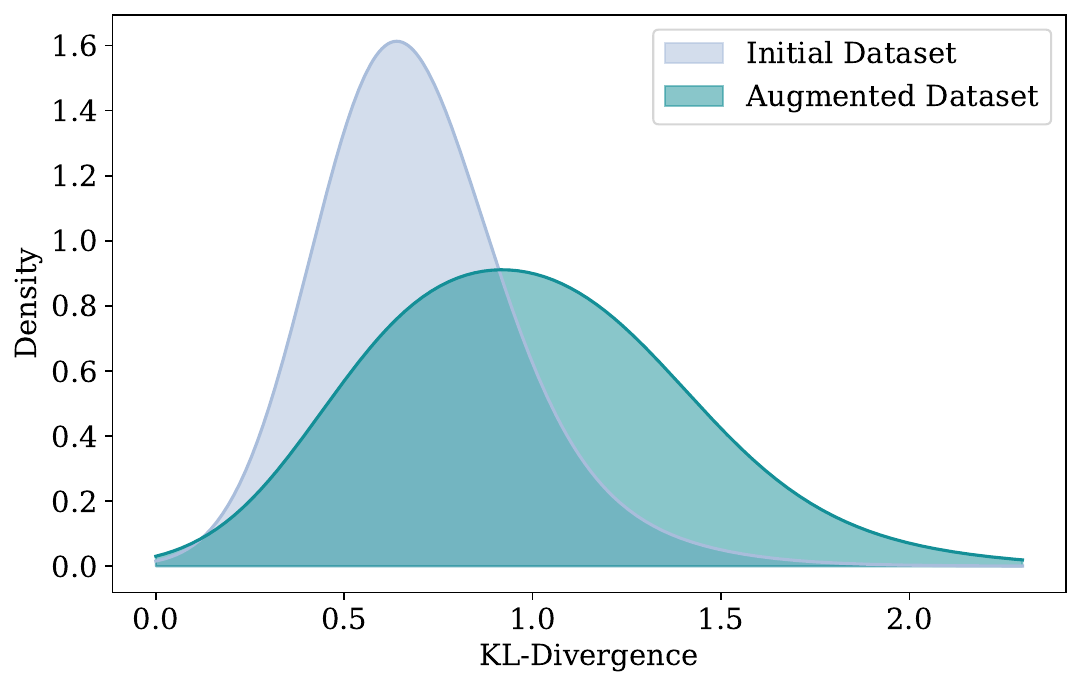}}
  \vspace{-2mm}
  \caption{(a) Visual Comparison of Mixup and Our Proposed Soft Attention-based Mixup. (b) Comparison of Probability Density Distribution of  the Initial and Augmented Datasets.}
  \vspace{-3mm}
\end{figure*}

\subsection{RoboMixup}
\label{sec:data_augment}

In order to address corruption and central bias issues, we design the RoboMixup as a novel robust data augmentation strategy, by generating augmentation pairs $(\tilde{I}$, $\tilde{Y})$ from image-pseudo labels pairs $(I$, $\hat{Y}$). 

\noindent \textbf{1) Design for Corruption Robustness.}~We propose an iterative data augmentation strategy that combines soft attention-based Mixup with dynamic augmentation to address the challenge of corruption robustness. To be detailed, the soft attention-based Mixup generates crucial and realistic samples, and dynamic augmentation selectively augments samples to achieve a more uniform data distribution, thereby enhancing the robustness of attention prediction models in self-driving scenarios by training with extra generated examples.

\textbf{Soft Attention-based Mixup.}~Given the training sample image and pseudo labels pair~$(I_i,\ \hat{Y}_i)$, $(I_j,\ \hat{Y}_j)$~,~the Mixup method generates new samples as follows: 
\begin{equation}
\tilde{I}=\ \lambda I_i+\left(1-\lambda\right)I_j;\ \tilde{Y}=\ \lambda \hat{Y}_i+\left(1-\lambda\right)\hat{Y}_j,
\label{eq:mixup}
\end{equation}
where $\lambda\in\left[0,\ 1\right]$, representing the transparencies used to blend the images and pseudo labels, typically assumed to follow a Beta distribution. Our method proposes using the attention map $S$ predicted by our model to replace $\lambda$ :
\begin{equation}
\begin{aligned}
\tilde{I} &= \left(\frac{S_i}{S_i + S_j} \odot I_i\right) + \left(\frac{S_j}{S_i + S_j} \odot I_j\right), \\
\tilde{Y} &= \left(\frac{S_i}{S_i + S_j} \odot \hat{Y}_i\right) + \left(\frac{S_j}{S_i + S_j} \odot \hat{Y}_j\right),
\end{aligned}
\label{eq:sa_mixup}
\end{equation}
where $\odot$ denotes pixel-wise multiplication. Rather than applying global transparency to the entire sample, our method employs pixel-level weights in sample mixing. This approach benefits from utilizing attention mechanisms to distinguish between critical objects in the scene, such as pedestrians and stop signs, and irrelevant backgrounds. On one hand, key objects are blended with higher attention weights in the mixup, increasing their presence and creating difficult samples; on the other hand, insignificant backgrounds are mixed with lower attention weights, rendering the samples more realistic. We show the generated examples of our method and traditional Mixup in Figure~\ref{fig:mixup_visual}. It can be observed that traditional Mixup overlays one image onto another as a ``ghost'' image. In contrast, our proposed method ``cuts out'' pedestrians, cyclists, and traffic lights from the first image and integrates them into the second image, thereby achieving a more realistic and complex image.  

\textbf{Dynamic Augmentation Strategy.}~
We further propose a dynamic augmentation strategy to alter the dataset distribution. Specifically, we use the KL divergence between the estimated attention map \( S_i \) and the average attention map \( S_{\text{avg}} \) as the selection criterion.~\( S_{\text{avg}} \) presents a Gaussian distribution centered around the road~\cite{xia2018predicting, palazzi2018predicting}. A higher KL divergence between \( S_i \) and \( S_{\text{avg}} \) indicates a lower similarity between the sample's attention distribution and the Gaussian distribution, signifying greater scene complexity. Utilizing this principle, we select the top \(K\)-percent samples within each batch with the highest KL divergence for applying Soft Attention-based Mixup, efficiently generating augmentation candidates. Additionally, we apply this criterion to identify and select samples that deviate significantly from the average attention map to further augment the dataset. As illustrated in Figure~\ref{fig:distribution_comp}, our approach effectively increases the proportion of challenging samples, resulting in a more uniform distribution toward the tail. The algorithm details can be found in the supplementary materials.



\noindent \textbf{2) Design for Central Bias.}~We propose a method that combines the advantages of random cropping and Mixup.

\noindent

\textbf{Random Crop.}  
Compared to directly resizing images to a fixed size, random cropping preserves the local granular details of the images. It also increases the likelihood of avoiding areas with significant central bias, thereby mitigating the inherent central bias in the dataset. After incorporating random crop data, we modify the model's training loss as follows:
\begin{equation}
\mathcal{L}_{\text{unc\_aug}} = \mathcal{L}_{\text{unc\_rcp}} + \mathcal{L}_{\text{unc}},
\label{eq:unc_aug}
\end{equation}
where \(\mathcal{L}_{\text{unc}}\) represents the loss calculated for the original data, as defined in Eq.~(\ref{eq:total_loss}), and \(\mathcal{L}_{\text{unc\_rcp}}\) represents the corresponding loss calculated for the random crop data.

\textbf{RegMixup.}  
Instead of using the vanilla Mixup, we follow the RegMixup~\cite{pinto2022regmixup} by introducing two key modifications to the original Mixup technique. First, it changes the parameter \(\alpha\) of the beta distribution in vanilla Mixup from 1 to 10. Second, the augmented samples are not replacements for the original samples but are treated as additional samples when calculating the loss. The RegMixup loss can be expressed as follows:
\begin{equation}
\mathcal{L}_{\text{reg}} = \mathcal{L} + \eta \mathcal{L}_{*},
\label{eq:regmixup}
\end{equation}
where \(\mathcal{L}\) is the loss for the original data, \(\mathcal{L}_{*}\) is the loss for the Mixup augmented data, and \(\eta\) is a hyperparameter (set to 1 base on~\cite{pinto2022regmixup}) that tunes the loss ratio. 

While our objective is to combine random crop data augmentation with RegMixup. Specifically, we replace \(\mathcal{L}\) in Eq.~(\ref{eq:regmixup}) with \(\mathcal{L}_{\text{unc\_aug}}\) from Eq.~(\ref{eq:unc_aug}), resulting in the following loss function:
\begin{equation}
\label{eq:rcpreg}
\mathcal{L}_{\text{unc\_reg}} = \mathcal{L}_{\text{unc\_rcp}} + \mathcal{L}_{\text{unc}} + \eta(\mathcal{L}_{\text{unc\_rcp}^*} + \mathcal{L}_{\text{unc}^*}),
\end{equation}
where \(\mathcal{L}_{\text{unc\_rcp}^*}\) represents the loss calculated for the random crop data after applying Mixup, and \(\mathcal{L}_{\text{unc}^*}\) denotes the loss calculated for the original data after applying Mixup.

\begin{figure*}
    \centering
    \includegraphics[width=1\linewidth]{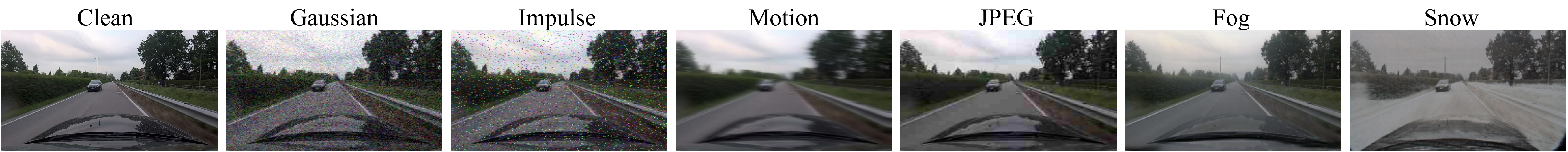}
    \vspace{-6mm}
    \caption{Visualization of our generated examples, highlighting how each corruption type substantially alters the original image representation.}
    \vspace{-2mm}
    \label{fig:corscene_vis}
\end{figure*}

\subsection{Optimization and Inference}
For optimization, we use the loss function \( \mathcal{L}_{unc} \) as defined in Eq.~(\ref{eq:total_loss}), and apply it to the dataset constructed using our proposed Soft Attention-based Mixup and the Dynamic Augmentation Strategy to overcome the corruption. In order to alleviate the central bias, we utilize the loss function \( \mathcal{L}_{unc\_reg} \) as shown in Eq.~(\ref{eq:rcpreg}). During the inference phase, we no longer use the uncertainty mining block and RoboMixup. Instead, we use only the attention prediction branch for evaluation.

\section{DriverAttention-C Dataset}
\label{sec:DriverAttention-C}

The real traffic scenarios often arise from camera input corruptions caused by various noise interferences and adverse weather conditions. However, no prior work has addressed this challenge in the context of self-driving attention prediction. To fill this gap, we collect the first \textit{DriverAttention-C} dataset, which consists of three subsets: \textit{BDD-A-C}, \textit{DR(eye)VE-C}, and \textit{DADA-2000-C}. Each subset contains four categories of common corruptions, \textit{i.e.,} \textit{noise}, \textit{blur}, \textit{digital}, and \textit{weather conditions}, which include six specific types of corruption: \textit{Gaussian Noise}, \textit{Impulse Noise}, \textit{Motion Blur}, \textit{JPEG Compression}, \textit{Fog}, and \textit{Snow}.

\subsection{Data Source}
Our dataset sources originate from test sets of BDD-A~\cite{xia2018predicting}, DR(eye)VE~\cite{alletto2016dr}, and DADA-2000~\cite{fang2021dada}. These datasets are derived from various collection driving contexts, including in-lab settings, in-car recordings, and real-world crash scenarios. We synthesize noisy corruption data based on these images, while we opt for \textit{CycleGAN}~\cite{CycleGAN2017} to generate realistic fog and snow effects. To achieve this, the target adverse weather domain training data are collected from the fog dataset \textit{SeeingThroughFog}~\cite{gated2depth2019}, the snow dataset \textit{WADS}~\cite{kurup2021dsor}, and the \textit{CADC}~\cite{kurup2021dsor}. As a result, our DriverAttetnion-C has a total of 115,332 images, which contains four subsets, \textit{i.e.,} BDD-A-C, DR(eye)VE-C, and DADA-2000-C, comprising 6,817, 7,497, and 4,908 test images, respectively.

\subsection{Generation Pipeline}

For corruptions data generation, we follow the open-source toolbox from Hendrycks et al.~\cite{hendrycks2019benchmarking} to simulate various degradation scenarios. These corruptions include Gaussian noise, impulse noise, motion blur, and JPEG compression, each chosen for their relevance to real-world challenges in image capture and transmission. Gaussian noise arises in low-illumination conditions or during prolonged camera operation, which may lead to sensor overheating. The noise is modeled by adding random Gaussian perturbations to the pixel values of the image. Impulse noise, also known as salt-and-pepper noise, typically results from sudden disturbances in signal transmission. This noise introduces random bright and dark pixels in the image, 6\% of the pixels are randomly affected. Motion blur occurs when the camera is unable to compensate for the rapid movement of the vehicle, resulting in smeared images. JPEG compression is a widely used lossy image compression method. Due to transmission constraints, aggressive compression can introduce artifacts such as blockiness and loss of fine details. 

Regarding adverse weather conditions like fog and snow, instead of manually simulating weather effects~\cite{hendrycks2019benchmarking}, we employ \textit{Cycle-GAN} to transfer the source domain, consisting of clear-weather images with good visibility, to the target domains of fog and snow. This approach ensures that the generated snow and fog images remain consistent with the underlying structure of the original clear-weather images while effectively simulating the target domain characteristics. Specifically, we use the BDD-A dataset as the source domain data and the aforementioned fog and snow datasets to train fog and snow generators, respectively, with each image having a resolution of \(512 \times 256\) pixels. Finally, we apply the trained generators to all three datasets to obtain the simulated images and filter out images with artificial stripes caused by inherent issues of the Cycle-GAN, resulting in 38,444 images. We show how each type of corruption significantly alters the original representation of the images in Figure~\ref{fig:corscene_vis}. More details about our dataset please refer to our supplementary.


\section{Experimental Results}

In experiments, we initially compare our unsupervised method against fully-supervised approaches across several well-established datasets: BDD-A, DR(eye)VE, DADA-2000 and DriverAttention-C. We then conduct thorough ablation studies to assess the effectiveness of each component.

\subsection{Experimental Settings}

\textbf{Datasets.} Our model's performance was assessed using three self-driving benchmarks: BDD-A, DR(eye)VE, and DADA-2000. \textbf{BDD-A}~\cite{xia2018predicting}, an in-lab driving attention dataset, comprises 1,232 short clips (each under 10 seconds) featuring diverse urban and rural road driving scenarios. Following this split, we use 28k frames for training, 6k for validation, and 9k for testing. \textbf{DR(eye)VE}~\cite{alletto2016dr}, an in-car dataset, aims for consistent driving conditions across its 74 videos (each up to 5 minutes). Following previous protocols~\cite{alletto2016dr}, we selected the last 37 videos for the test set. \textbf{DADA-2000}~\cite{fang2021dada}, including vehicle crash scenarios, allows for the prediction of driving attention under critical conditions. With over 658,746 frames across 2000 clips, we adopt a 3:1:1 split ratio for training, validation, and testing as per standard practices~\cite{fang2021dada}. To evaluate the robustness of the self-driving attention prediction model, we test them on our constructed \textbf{DRIVERATTENTION-C} dataset, as described in Section~\ref{sec:DriverAttention-C}. To evaluate the models' generalization capabilities to corrupted data, all methods are trained solely on clean datasets and tested on the corrupted datasets. 

\begin{table*}
\centering
\caption{Performance comparison between our proposed unsupervised method and state-of-the-art fully-supervised methods. The numbers in bold denote the best results, and those marked with underlines denote the second best.}
\vspace{-3mm}
\label{table:overview}
\setlength\tabcolsep{2.5mm}
\begin{tabular}{lcccccc}
\toprule
\multicolumn{1}{c}{\textbf{Methods}} &
  \multicolumn{2}{|c}{\textbf{BDD-A}~\cite{xia2018predicting}} &
  \multicolumn{2}{|c}{\textbf{DR(eye)VE}~\cite{alletto2016dr}} &
  \multicolumn{2}{|c}{\textbf{DADA-2000~\cite{fang2021dada}}} \\ 
\multicolumn{1}{c}{} &
  \multicolumn{1}{|c}{KLD↓} &
  \multicolumn{1}{c|}{CC↑} &
  KLD↓ &
  \multicolumn{1}{c|}{CC↑} &
  KLD↓ &
  CC↑ \\ 
\midrule\midrule
\multicolumn{1}{l|}{Multi-Branch~\cite{palazzi2018predicting}} &
  \underline{1.23} &
  \multicolumn{1}{c|}{0.57} &
  {\textbf{1.84}} &
  \multicolumn{1}{c|}{{\textbf{0.52}}} &
  1.90 &
  0.41 \\
\multicolumn{1}{l|}{HWS~\cite{xia2018predicting}} &
  1.27 &
  \multicolumn{1}{c|}{0.57} &
  2.08 &
  \multicolumn{1}{c|}{{0.47}} &
  \underline{1.87} &
  0.42 \\
\multicolumn{1}{l|}{SAM~\cite{cornia2018predicting}} &
  1.91 &
  \multicolumn{1}{c|}{0.37} &
  2.38 &
  \multicolumn{1}{c|}{0.37} &
  2.45 &
  0.28 \\
\multicolumn{1}{l|}{Tased-Net~\cite{min2019tased}} &
  1.52 &
  \multicolumn{1}{c|}{0.54} &
  {2.24} &
  \multicolumn{1}{c|}{0.49} &
  {1.98} &
  {0.42} \\
\multicolumn{1}{l|}{MEDIRL~\cite{baee2021medirl}} &
  2.51 &
  \multicolumn{1}{c|}{{\textbf{0.74}}} &
  - &
  \multicolumn{1}{c|}{-} &
  2.93 &
  {\textbf{0.63}} \\
\multicolumn{1}{l|}{ML-Net~\cite{cornia2016deep}} &
  2.25 &
    \multicolumn{1}{c|}{{0.26}} &
  3.03 &
  \multicolumn{1}{c|}{0.20} &
  2.71 &
  0.21 \\
\multicolumn{1}{l|}{UNISAL~\cite{droste2020unified}} &
  1.64 &
  \multicolumn{1}{c|}{0.44} &
  2.51 &
  \multicolumn{1}{c|}{0.34} &
  2.31 &
  0.31 \\
\multicolumn{1}{l|}{PiCANet~\cite{liu2018picanet}} &
  {1.43} &
  \multicolumn{1}{c|}{{0.49}} &
  2.03 &
  \multicolumn{1}{c|}{0.47} &
  1.95 &
  0.40 \\
\multicolumn{1}{l|}{DADA~\cite{fang2021dada}} &
  1.55 &
  \multicolumn{1}{c|}{0.55} &
  1.97 &
  \multicolumn{1}{c|}{0.49} &
  2.24 &
  0.40 \\ 
\rowcolor{gray!20}
\multicolumn{1}{l|}{\textbf{Ours (unsupervised)}} &
  {\textbf{1.099±0.016}} &
  \multicolumn{1}{c|}{\underline{0.640±0.007}} &
  \underline{1.901±0.004} &
  \multicolumn{1}{c|}{{\underline{0.510±0.005}}} &
  {\textbf{1.677±0.007}} &
  \underline{0.488±0.002} \\ 
  \bottomrule
\end{tabular}
\vspace{-3mm}
\end{table*}

\textbf{Metrics.} We employs two prevalent metrics: Kullback-Leibler divergence (KLD)~\cite{cornia2016deep} and Pearson Correlation Coefficient (CC~\cite{palazzi2018predicting}). KLD measures the similarity between the predicted and actual driving attention distributions. This asymmetric measure penalizes false negatives more severely than false positives. CC, on the other hand, assesses the linear correlation between the predicted and actual distributions, equally penalizing both false negatives and false positives in a symmetric manner. We refrain from using discrete metrics like Area Under ROC Curve ( ($\textit{AUC}$)) and its variants($\textit{AUC-J}, \textit{AUC-S}$), Normalized Scanpath Saliency ($\textit{NSS}$), and Information Gain ($\textit{IG}$)~\cite{bylinskii2018different}, opting instead for continuous distribution metrics better suited for identifying risk-related pixels and areas in driving contexts~\cite{pal2020looking}.

In order to compare the robustness of various methods, inspired by~\cite{hendrycks2019benchmarking}, we propose two new metrics based on the ratio to the baseline ($ref$). We introduce the term Degradation (D) to denote the increase in error rate attributable to corruption.  For the Kullback-Leibler divergence (KLD), $D_{kld} = KLD$, and for the Pearson Correlation Coefficient (CC), $D_{cc} = 1 - CC$. Then we introduce the mean Corruption Degradation (mCD) metric for all types of corruptions $C\ =\{Gaussian, Impulse, Motion, JPEG, Fog, Snow\}$ in a given model $f$, given by:
\begin{equation}
\text{mCD}^f = \sum_{c \in C} D_c^f / \sum_{c \in C} D_c^{ref} .
\label{eq:mcd}
\end{equation}

We further introduce the Relative mean Corruption Degradation (Relative mCD) metric to capture the discrepancy between clean data and corrupted data, which is given by: 
\begin{equation}
\text{Relative mCD}^f = \sum_{c \in C} {(D_c^f-D_{clean}^f)} / \sum_{c \in C} {(D_c^{ref}-D_{clean}^{ref})}.
\label{eq:mcd}
\end{equation}

Here We select MLNet~\cite{cornia2016deep} as the baseline and report both metrics in experiments.



\noindent\textbf{Compared Methods.} We compare our proposed unsupervised approach with recent fully-supervised state-of-the-art methods, including Multi-Branch~\cite{palazzi2018predicting}, HWS~\cite{xia2018predicting}, DADA~\cite{fang2021dada}, and MEDIRL~\cite{baee2021medirl} trained on the corresponding autonomous driving datasets; and SAM~\cite{cornia2018predicting}, TASED-Net~\cite{min2019tased}, ML-Net~\cite{cornia2016deep}, UNISAL~\cite{droste2020unified}, and PiCANet~\cite{liu2018picanet} trained on natural scenes. To evaluate robustness, we compared our method with state-of-the-art data augmentation techniques, including Mixup~\cite{zhang2018mixup}, RegMixup~\cite{pinto2022regmixup}, TransMix~\cite{chen2022transmix}, Random Crop~\cite{palazzi2018predicting}, and Human Weighted Sampling~\cite{xia2018predicting}. Note that we denote our method in conferce paper as UAP, and our method with RoboMixup as RUAP in this manuscript.

\subsection{Implementation Details}
We implement our network using PyTorch~\cite{paszke2019pytorch}. For each dataset, video frames and gaze annotated maps are sampled at 3Hz to ensure alignment. During training, pseudo-labels and original images are resized to $224 \times 224$, with values normalized spatially. For knowledge embedding, Mask R-CNN pre-trained on MS-COCO~\cite{lin2014microsoft} segments key instances for fusion with pseudo-labels. In Section~\ref{sec:kms}, we set the coverage threshold \( p \) to 98 and the proportion factor \( \eta \) to 0.1, and the adjustment factor \( \alpha \) to 0.3. In the dynamic augmentation strategy from Section~\ref{sec:data_augment}, the top \( K \)-percent of samples within each batch is set to \( 1/8 \). The initial learning rate is set to $0.001$, with a scheduler that warms up before descending cosinely. During training, we run the model for $10$ epochs with a batch size of $32$, taking approximately 50 minutes on an RTX 3090 GPU. Inference of attention regions per frame takes about $12$~ms. Please refer to the supplementary for more details. 


\subsection{Quantitative Comparisons}

Table~\ref{table:overview} presents the quantitative performance comparison between our unsupervised network and fully-supervised state-of-the-art models. It is noteworthy that our model, trained exclusively on pseudo-labels from the BDD-A dataset without using any ground-truth labels, was tested across various benchmarks. Table~\ref{table:overview} shows that our proposed network not only competes with but also surpasses many fully-supervised methods w.r.t KLD scores on BDD-A and DADA-2000, and ranks second in CC on BDD-A and DR(eye)VE. This underscores the effectiveness and potential of our unsupervised approach. Moreover, to assess the transferability across three self-driving benchmarks (BDD-A, DR(eye)VE, DADA-2000), we present the results of training our method with pseudo-labels from each dataset and testing on the others in Table~\ref{table:three}. The model, when trained on pseudo-labels from BDD-A, achieves superior performance on the test sets of DADA-2000 and itself. For the DR(eye)VE test set, the network trained on DR(eye)VE pseudo-labels excels in CC, and the one trained on DADA-2000 pseudo-labels leads in KLD, showcasing our method's robust transferability. Because of BDD-A images depicting a wide range of driving scenarios, our model employs pseudo-labels generated from BDD-A.

\begin{table*}
\centering
\caption{Performance comparison of our proposed unsupervised network trained with pseudo-labels generated from various self-driving datasets (BDD-A, DR(eye)VE, DADA-2000) and then test on each benchmark. The best result is highlighted in bold.}
\vspace{-3mm}
\label{table:three}
\setlength\tabcolsep{1.7mm}
\begin{tabular}{c|cc|cc|cc}
\toprule
\multicolumn{1}{c}{\textbf{Pseudo-labels}} &
  \multicolumn{2}{|c}{BDD-A~\cite{xia2018predicting}} &
  \multicolumn{2}{|c}{DR(eye)VE~\cite{alletto2016dr}} &
  \multicolumn{2}{|c}{DADA-2000~\cite{fang2021dada}} \\ 
\multicolumn{1}{c}{} &
  \multicolumn{1}{|c}{KLD↓} &
  CC↑ &
  \multicolumn{1}{|c}{KLD↓} &
  CC↑ &
  \multicolumn{1}{|c}{KLD↓} &
  CC↑ \\ 
  \midrule\midrule
BDD-A             & \textbf{1.099±0.016} & \textbf{0.635±0.007} & 1.924±0.004 & 0.508±0.003 & \textbf{1.677±0.007} & \textbf{0.488±0.002} \\
DR(eye)VE        & 1.188±0.011 & 0.608±0.002 & 1.908±0.008 & \textbf{0.517±0.005} & 1.801±0.017 & 0.458±0.004 \\
DADA-2000        & 1.242±0.021 & 0.578±0.009 & \textbf{1.889±0.012} & 0.513±0.010 & 1.711±0.015 & 0.483±0.007 \\ \bottomrule
\end{tabular}
\vspace{-2mm}
\end{table*}

\begin{table*}[]
\centering
\caption{Performance comparison of our proposed unsupervised model with state-of-the-art fully-supervised methods on clean and various noisy conditions. The mCD value represents the mean Corruption Degradation across the Noise, Blur, Digital, and Weather categories. The numbers in bold denote the best results, while those underlined indicate the second-best.}
\vspace{-3mm}
\label{table:noise_robo_comp}
\resizebox{\textwidth}{!}{%
\begin{tabular}{@{}l|cc|cc|cc|cc|cc|cc|cc|cc|cccc@{}}
\toprule
\textbf{Methods} &
  \multicolumn{2}{c|}{\textbf{Clean}} &
  \multicolumn{2}{c|}{\textbf{Gaussian}} &
  \multicolumn{2}{c|}{\textbf{Impulse}} &
  \multicolumn{2}{c|}{\textbf{Motion}} &
  \multicolumn{2}{c|}{\textbf{JPEG}} &
  \multicolumn{2}{c|}{\textbf{Fog}} &
  \multicolumn{2}{c|}{\textbf{Snow}} &
  \multicolumn{2}{c|}{\textbf{mCD}} &
  \multicolumn{2}{c}{\textbf{Relative mCD}}   \\
 & KLD↓ & CC↑ & KLD↓ & CC↑ & KLD↓ & CC↑ & KLD↓ & CC↑ & KLD↓ & CC↑ & KLD↓ & CC↑ & KLD↓ & CC↑ & KLD↓ & CC↓ & KLD↓ & CC↓ \\ \midrule
\multicolumn{19}{c}{\textbf{BDD-A-C}} \\ \midrule
Multi-Branch~\cite{palazzi2018predicting}         & 1.225 & 0.570 & 2.426 & 0.173 & 2.072 & 0.291 & 1.636 & 0.428 & 1.533 & 0.459 & 1.290 & 0.553 & 1.469 & 0.488 & 0.736 & 0.754 & 3.693 & 3.779 \\
HWS~\cite{xia2018predicting}                 & 1.263 & 0.568 & 1.721 & 0.404 & 1.837 & 0.369 & 1.498 & 0.482 & 1.303 & 0.553 & 1.276 & 0.563 & 1.602 & 0.439 & 0.665 & 0.677 & 1.992 & 2.199 \\
SAM~\cite{cornia2018predicting}                 & 1.604 & 0.452 & 1.790 & 0.389 & 1.859 & 0.377 & 1.750 & 0.394 & 1.707 & 0.416 & 1.597 & 0.454 & 1.894 & 0.358 & 0.763 & 0.767 & 1.168 & 1.191 \\
TASED-NET~\cite{min2019tased}    & 1.558 & 0.533 & 1.839 & 0.453 & 1.902 & 0.444 & 1.808 & 0.454 & 1.627 & 0.504 & 1.558 & 0.535 & 1.723 & 0.469 & 0.753 & 0.667 & 1.331 & 1.246 \\
ML-Net~\cite{cornia2016deep}             & 2.177 & 0.260 & 2.306 & 0.207 & 2.424 & 0.184 & 2.324 & 0.208 & 2.263 & 0.226 & 2.191 & 0.255 & 2.387 & 0.208 & 1.000 & 1.000 & 1.000 & \underline{1.000} \\
UNISAL~\cite{droste2020unified}              & 1.639 & 0.440 & 1.891 & 0.347 & 1.879 & 0.356 & 2.003 & 0.301 & 1.826 & 0.369 & 1.677 & 0.426 & 1.861 & 0.361 & 0.802 & 0.815 & 1.564 & 1.765 \\
PiCaNet~\cite{liu2018picanet}           & 1.432 & 0.490 & 1.741 & 0.395 & 1.715 & 0.402 & 1.910 & 0.327 & 1.586 & 0.435 & 1.475 & 0.475 & 1.694 & 0.396 & 0.728 & 0.758 & 1.836 & 1.875 \\
DADA~\cite{fang2021dada}               & 1.526 & 0.551 & \underline{1.555} & \underline{0.501} & 1.658 & 0.464 & 1.639 & 0.498 & 1.539 & 0.537 & 1.543 & 0.546 & 1.734 & 0.477 & 0.696 & 0.632 & \textbf{0.615} & 1.040 \\
\rowcolor{gray!20}
UAP(unsupervised)       & \textbf{1.112} & \underline{0.627} & 1.568 & 0.486 & \underline{1.625} & \underline{0.465} & \underline{1.330} & \underline{0.555} & \underline{1.246} & \underline{0.582} & \textbf{1.130} & \underline{0.623} & \underline{1.300} & \underline{0.564} & \underline{0.590} & \underline{0.578} & 1.833 & 1.790 \\
\rowcolor{gray!20}
RUAP(unsupervised)    & \underline{1.121} & \textbf{0.628} & \textbf{1.294} & \textbf{0.563} & \textbf{1.321} & \textbf{0.559} & \textbf{1.194} & \textbf{0.599} & \textbf{1.167} & \textbf{0.611} & \underline{1.131} & \textbf{0.626} & \textbf{1.248} & \textbf{0.580} & \textbf{0.529} & \textbf{0.522} & \underline{0.755} & \textbf{0.846} \\ \midrule
\multicolumn{19}{c}{\textbf{DR(eye)VE-C}} \\ \midrule
Multi-Branch~\cite{palazzi2018predicting}         & \textbf{1.806} & \underline{0.505} & \textbf{1.868} & \textbf{0.501} & \textbf{1.901} & \textbf{0.495} & \textbf{1.833} & \textbf{0.504} & \textbf{1.804} & \textbf{0.508} & \textbf{1.855} & \textbf{0.502} & 2.085 & 0.445 & \textbf{0.598} & \textbf{0.592} & 0.505 & \textbf{0.314} \\
HWS~\cite{xia2018predicting}                          & 2.116 & 0.443 & 2.719 & 0.319 & 2.979 & 0.242 & 2.614 & 0.350 & 2.177 & 0.432 & 2.192 & 0.427 & 2.392 & 0.349 & 0.794 & 0.754 & 2.356 & 2.255 \\
SAM~\cite{cornia2018predicting}                          & 2.301 & 0.383 & 2.526 & 0.312 & 2.549 & 0.317 & 2.509 & 0.315 & 2.446 & 0.338 & 2.356 & 0.364 & 2.534 & 0.313 & 0.786 & 0.785 & 1.104 & 1.418 \\
TASED-NET~\cite{min2019tased}              & 2.303 & 0.482 & 2.500 & 0.430 & 2.448 & 0.430 & 2.484 & 0.431 & 2.365 & 0.469 & 2.315 & 0.472 & 2.358 & 0.463 & 0.763 & 0.642 & 0.646 & 0.824 \\
ML-Net~\cite{cornia2016deep}                      & 2.994 & 0.182 & 3.067 & 0.157 & 3.222 & 0.130 & 3.216 & 0.124 & 3.093 & 0.155 & 3.125 & 0.152 & 3.250 & 0.135 & 1.000 & 1.000 & 1.000 & 1.000 \\
UNISAL~\cite{droste2020unified}                       & 2.425 & 0.333 & 2.569 & 0.279 & 2.596 & 0.283 & 2.740 & 0.234 & 2.576 & 0.284 & 2.476 & 0.320 & 2.532 & 0.301 & 0.816 & 0.835 & 0.931 & 1.243 \\
PiCaNet~\cite{liu2018picanet}                    & 2.049 & 0.438 & 2.167 & 0.412 & 2.191 & 0.415 & 2.208 & 0.385 & 2.079 & 0.429 & 2.027 & 0.441 & 2.074 & 0.422 & 0.672 & 0.679 & \underline{0.448} & 0.519 \\
DADA~\cite{fang2021dada}                        & 2.125 & 0.454 & 2.139 & 0.446 & 2.164 & 0.438 & 2.127 & 0.452 & 2.125 & 0.460 & 2.107 & 0.454 & 2.300 & 0.393 & 0.683 & 0.652 & \textbf{0.210} & \underline{0.339} \\
\rowcolor{gray!20}
UAP(unsupervised)       & 1.882 & 0.487 & 2.402 & 0.362 & 2.489 & 0.331 & 2.016 & 0.464 & 1.982 & 0.468 & 1.894 & 0.490 & \underline{1.985} & \underline{0.471} & 0.673 & 0.663 & 1.463 & 1.406 \\
\rowcolor{gray!20}
RUAP(unsupervised)    & \underline{1.825} & \textbf{0.507} & \underline{1.988} & \underline{0.467} & \underline{2.035} & \underline{0.458} & \underline{1.891} & \underline{0.491} & \underline{1.884} & \underline{0.494} & \underline{1.871} & \underline{0.497} & \textbf{1.939} & \textbf{0.486} & \underline{0.612} & \underline{0.604} & 0.652 & 0.623  \\ \midrule
\multicolumn{19}{c}{\textbf{DADA-2000-C}} \\ \midrule
Multi-Branch~\cite{palazzi2018predicting}         & 1.905 & 0.413 & 2.660 & 0.225 & 2.449 & 0.286 & 2.529 & 0.260 & 2.154 & 0.343 & 1.982 & 0.390 & 2.056 & 0.367 & 0.792 & 0.824 & 3.221 & 3.246 \\
HWS~\cite{xia2018predicting}                          & 1.882 & 0.420 & 2.407 & 0.295 & 2.527 & 0.264 & 2.007 & 0.385 & 1.885 & 0.418 & 1.890 & 0.417 & 2.174 & 0.345 & 0.738 & 0.773 & 2.145 & 2.118 \\
SAM~\cite{cornia2018predicting}                          & 2.290 & 0.316 & 2.340 & 0.301 & 2.402 & 0.297 & 2.357 & 0.293 & 2.322 & 0.306 & 2.380 & 0.288 & 2.475 & 0.260 & 0.817 & 0.849 & \underline{0.719} & 0.807 \\
TASED-NET~\cite{min2019tased}              & 2.002 & 0.412 & 2.285 & 0.367 & 2.312 & 0.362 & 2.192 & 0.376 & 2.067 & 0.399 & 1.967 & 0.417 & 2.008 & 0.401 & 0.735 & 0.734 & 1.099 & \underline{0.802} \\
ML-Net~\cite{cornia2016deep}                      & 2.787 & 0.196 & 2.854 & 0.170 & 3.019 & 0.143 & 2.870 & 0.172 & 2.826 & 0.183 & 2.860 & 0.178 & 3.038 & 0.143 & 1.000 & 1.000 & 1.000 & 1.000 \\
UNISAL~\cite{droste2020unified}                       & 2.310 & 0.312 & 2.410 & 0.276 & 2.430 & 0.279 & 2.505 & 0.251 & 2.408 & 0.283 & 2.318 & 0.307 & 2.516 & 0.267 & 0.835 & 0.865 & 0.976 & 1.118 \\
PiCaNet~\cite{liu2018picanet}                    & 1.941 & 0.401 & 2.240 & 0.331 & 2.215 & 0.338 & 2.329 & 0.290 & 2.065 & 0.365 & 1.967 & 0.391 & 2.098 & 0.353 & 0.739 & 0.785 & 1.702 & 1.807 \\
DADA~\cite{fang2021dada}                        & 2.240 & 0.399 & \underline{2.171} & \underline{0.401} & \underline{2.177} & \underline{0.397} & 2.252 & 0.380 & 2.242 & 0.395 & 2.234 & 0.398 & 2.397 & 0.348 & 0.771 & 0.735 & \textbf{0.044} & \textbf{0.401} \\
\rowcolor{gray!20}
UAP(unsupervised)       & \underline{1.735} & \textbf{0.483} & 2.345 & 0.328 & 2.381 & 0.322 & \underline{1.954} & \underline{0.432} & \underline{1.829} & \underline{0.464} & \underline{1.745} & \underline{0.480} & \underline{1.856} & \underline{0.446} & \underline{0.693} & \underline{0.704} & 2.282 & 2.278 \\
\rowcolor{gray!20}
RUAP(unsupervised)    & \textbf{1.727} & \textbf{0.483} & \textbf{1.998} & \textbf{0.421} & \textbf{2.047} & \textbf{0.413} & \textbf{1.820} & \textbf{0.461} & \textbf{1.757} & \textbf{0.476} & \textbf{1.733} & \textbf{0.483} & \textbf{1.854} & \textbf{0.448} & \textbf{0.642} & \textbf{0.658} & 1.137 & 1.048 \\
\bottomrule
\end{tabular}}
\vspace{-1mm}
\end{table*}

\begin{table*}[ht!]
\centering
\caption{Robustness comparison of our proposed RoboMixup with other data augmentation techniques on the BDD-A-C, using our UAP as the baseline. The bold numbers represent the best results, while the underlined numbers indicate the second-best performance.}
\vspace{-3mm}
\label{table:corruption_mixup_comp}
\resizebox{\textwidth}{!}{%
\begin{tabular}{@{}l|cc|cc|cc|cc|cc|cc|cc|cc|cccc@{}}
\toprule
\textbf{Methods} &
  \multicolumn{2}{c|}{\textbf{Clean}} &
  \multicolumn{2}{c|}{\textbf{Gaussian Noise}} &
  \multicolumn{2}{c|}{\textbf{Impulse Noise}} &
  \multicolumn{2}{c|}{\textbf{Motion Blur}} &
  \multicolumn{2}{c|}{\textbf{JPEG Compression}} &
  \multicolumn{2}{c|}{\textbf{Fog}} &
  \multicolumn{2}{c|}{\textbf{Snow}} &
  \multicolumn{2}{c|}{\textbf{mCD}} &
  \multicolumn{2}{c}{\textbf{Relative mCD}}   \\
 & KLD$\downarrow$ & CC$\uparrow$ & KLD$\downarrow$ & CC$\uparrow$ & KLD$\downarrow$ & CC$\uparrow$ & KLD$\downarrow$ & CC$\uparrow$ & KLD$\downarrow$ & CC$\uparrow$ & KLD$\downarrow$ & CC$\uparrow$ & KLD$\downarrow$ & CC$\uparrow$ & KLD$\downarrow$ & CC$\downarrow$ & KLD$\downarrow$ & CC$\downarrow$ \\
\midrule
Baseline &
  \textbf{1.112} & \underline{0.627} &
  1.568 & 0.486 &
  1.625 & 0.465 &
  1.330 & 0.555 &
  1.246 & 0.582 &
  \textbf{1.130} & \underline{0.623} &
  1.300 & 0.564 &
  1.000 & 1.000 &
  1.000 & 1.000 \\
Mixup~\cite{zhang2018mixup} &
  \underline{1.121} & 0.625 &
  \underline{1.388} & \underline{0.536} &
  \underline{1.408} & \underline{0.534} &
  1.437 & 0.536 &
  \underline{1.187} & \underline{0.606} &
  1.162 & 0.616 &
  1.419 & 0.550 &
  0.976 & 0.962 & 
  0.835 & 0.764\\
RegMixup~\cite{pinto2022regmixup} &
  1.334 & 0.593 &
  2.323 & 0.264 &
  2.467 & 0.203 &
  1.594 & 0.536 &
  1.451 & 0.571 &
  1.360 & 0.590 &
  1.469 & 0.567 &
  1.301 & 1.200 & 
  1.742 & 1.698\\
TransMix~\cite{chen2022transmix} &
  1.141 & 0.622 &
  1.447 & 0.515 &
  1.414 & 0.524 &
  \underline{1.302} & \underline{0.567} &
  1.195 & 0.604 &
  1.153 & 0.618 &
  \underline{1.249} & \textbf{0.581} &
  \underline{0.946} & \underline{0.951} & 
  \underline{0.599} & \underline{0.663}\\
\rowcolor{gray!20}
RoboMixup (ours) &
  \underline{1.121} & \textbf{0.628} &
  \textbf{1.294} & \textbf{0.563} &
  \textbf{1.321} & \textbf{0.559} &
  \textbf{1.194} & \textbf{0.599} &
  \textbf{1.167} & \textbf{0.611} &
  \underline{1.131} & \textbf{0.626} &
  \textbf{1.248} & \underline{0.580} &
  \textbf{0.897} & \textbf{0.903} & 
  \textbf{0.412} & \textbf{0.472} \\
\bottomrule
\end{tabular}}
\vspace{-2mm}
\end{table*}

\begin{table*}[ht]
\centering
\caption{Performance comparison of our proposed RoboMixup with other data augmentation techniques in terms of central bias on BDD-A, using our UAP as the baseline. The deviation threshold \(\delta\) indicates the severity of the central bias issue. The bold numbers represent the best results, while the underlined numbers indicate the second-best performance.}
\vspace{-2mm}
\label{table:central_bias_comp}
\begin{tabular}{l|cc|cc|cc|cc|cccccccccc}
\toprule
\textbf{Methods} & \multicolumn{2}{c|}{\boldmath$\delta=2.0$} & \multicolumn{2}{c|}{\boldmath$\delta=2.5$} & \multicolumn{2}{c|}{\boldmath$\delta=3.0$} & \multicolumn{2}{c|}{\boldmath$\delta=3.5$} & \multicolumn{2}{c}{\boldmath$\delta=4.0$} \\
                  & KLD$\downarrow$ & CC$\uparrow$ & KLD$\downarrow$ & CC$\uparrow$ & KLD$\downarrow$ & CC$\uparrow$ & KLD$\downarrow$ & CC$\uparrow$ & KLD$\downarrow$ & CC$\uparrow$ \\
\midrule
Baseline~\cite{zhu2023unsupervised}               & \underline{1.138}           & 0.622        & 1.217           & 0.598        & 1.318           & 0.564        & 1.435           & 0.523        & 1.559           & 0.484        \\
Random Crop~\cite{palazzi2018predicting}       & 1.182           & 0.619        & 1.243           & 0.598        & 1.319           & \underline{0.568}        & \underline{1.399}           & \underline{0.534}        & \underline{1.480}           & 0.503        \\
Human Weighted Sampling~\cite{xia2018predicting}          & \textbf{1.125}           & \underline{0.625}        & \underline{1.203}           & \underline{0.602}        & \underline{1.301}           & \underline{0.568}        & 1.417           & 0.527        & 1.542           & 0.488        \\
Mixup~\cite{zhang2018mixup}          & 1.152           & 0.619        & 1.237           & 0.593        & 1.347           & 0.555        & 1.467           & 0.512        & 1.598           &  0.470        \\
RegMixup~\cite{pinto2022regmixup}          & 1.347           & 0.592        & 1.404           & 0.576        & 1.466           & 0.555        & 1.524           & 0.531        & 1.582           & \underline{0.510}        \\
TransMix~\cite{chen2022transmix}          & 1.164           & 0.618        & 1.254           & 0.592        & 1.368           & 0.555        & 1.492           & 0.513        & 1.629           & 0.471        \\
\rowcolor{gray!20}
RoboMixup (ours)              & 1.139           & \textbf{0.629}        & \textbf{1.188}           & \textbf{0.616}        & \textbf{1.250}           & \textbf{0.593}        & \textbf{1.314}           & \textbf{0.568}        & \textbf{1.382}           & \textbf{0.544}        \\
\bottomrule
\end{tabular}
\vspace{-3mm}
\end{table*}

\begin{table*}
\centering
\caption{Comparison of our unsupervised model against its ablated versions. All models train with pseudo-labels from BDD-A and test across BDD-A, DR(eye)VE, and DADA-2000. Each iteration ablates components of the model until only the basic APB remains. The basic APB undergoes unsupervised training with ML-Net-generated pseudo-labels from the BDD-A training set. Best results are highlighted in bold.}
\vspace{-2mm}
\label{table:ablated}
\setlength\tabcolsep{4.6mm}
\begin{tabular}{l|ll|ll|ll}
\toprule
\multicolumn{1}{c|}{\textbf{Ablated Variants}} &
  \multicolumn{2}{c|}{\textbf{BDD-A}~\cite{xia2018predicting}} &
  \multicolumn{2}{c|}{\textbf{DR(eye)VE}~\cite{alletto2016dr}} &
  \multicolumn{2}{c}{\textbf{DADA-2000~\cite{fang2021dada}}} \\
\multicolumn{1}{c|}{} &
  KLD↓ &
  \multicolumn{1}{c|}{CC↑} &
  KLD↓ &
  \multicolumn{1}{c|}{CC↑} &
  KLD↓ &
  CC↑ \\ \midrule\midrule
APB~(unsupervised)                & 1.233 & 0.608 & 2.013 & 0.501 & 1.805 & 0.460 \\
APB+UMB                          & 1.141 & 0.622 & 1.941 & 0.510 & 1.702 & 0.480 \\
APB+UMB+non-local block          & 1.134 & 0.626 & 1.917 & 0.514 & 1.695 & 0.485 \\
\textbf{Ours}:~APB+UMB+non-local block+KEB & \textbf{1.099} & \textbf{0.635} & \textbf{1.901} & \textbf{0.518} & \textbf{1.677} & \textbf{0.488} \\ \bottomrule
\end{tabular}
\vspace{-2mm}
\end{table*}
\textbf{Corruption Robustness.}
\noindent{\bf 1). Benchmark.}~We first evaluated the model's performance under various types of corruption, analyzing its quantitative results using the KLD and CC metrics. As shown in Table~\ref{table:noise_robo_comp}, RUAP, trained using our data augmentation method proposed in Section~\ref{sec:data_augment}, demonstrated enhanced robustness compared to UAP and other fully-supervised methods across multiple datasets and corruption types, achieving a leading position.
Using ML-Net as a baseline, we assessed various models based on their mean Correlation Distance (mCD) and Relative mCD. As shown in Table~\ref{table:noise_robo_comp}, our RUAP model outperforms others in the mCD metric across all three benchmark tests, demonstrating its high robustness and excellent performance in predicting driver attention under various corruption conditions. Although RUAP achieves the best Relative mCD on the BDDA-C dataset in terms of KLD and performs moderately on DR(eye)VE-C and DADA-2000-C, we emphasize that other methods that perform best in this metric may indicate overall stable but poor performance across all scenarios (e.g., DADA), which diminishes their practical value. Overall, both metrics demonstrate that our proposed method excels in both performance and stability.~{\bf 2). Comparison with Other Data Augmentation Methods.}~As shown in Table~\ref{table:corruption_mixup_comp}, we compared our proposed RoboMixup with state-of-the-art data augmentation techniques on the BDD-A-C dataset, using UAP as the baseline. Our proposed method demonstrates the highest robustness across various corruptions. Specifically, our method reduces the (Relative) mCD values to 41.2\%/47.2\% compared to UAP and decreases mCD values to 89.7\%/90.3\% in terms of KLD and CC metrics, respectively. Besides, it is worth noting that our method outperforms the TransMix method, which also incorporates attention into Mixup but only at the instance level for labels. This highlights the effectiveness of our proposed approach, which utilizes attention at the pixel level combined with a dynamic augmentation strategy.

\textbf{Robustness against central bias.}~According to~\cite{xia2018predicting} which utilized the KL divergence with a specific threshold to select samples exhibiting less central bias as a test set, we extend it by introducing a set of deviation thresholds \( \delta \in \{2.0, 2.5, 3.0, 3.5, 4.0\} \) and selecting samples by calculating the KL divergence between the average attention map and individual attention maps that exceed the threshold. In Table~\ref{table:central_bias_comp}, we use UAP as the baseline to compare the performance of the proposed RoboMixup with other data augmentation methods on BDD-A. It can be observed that as the central bias problem becomes more severe (i.e., as the deviation threshold \(\delta\) increases), the greater the robustness of the proposed RoboMixup method. Specifically, compared to the baseline, the CC metric improves by 3.0\%, 5.1\%, 8.6\%, and 12.0\% at \(\delta = 2.5\), \(3.0\), \(3.5\), and \(4.0\), respectively. 

\subsection{Qualitative Results}
\begin{figure*}
    \centering
    \includegraphics[width=0.93\linewidth]{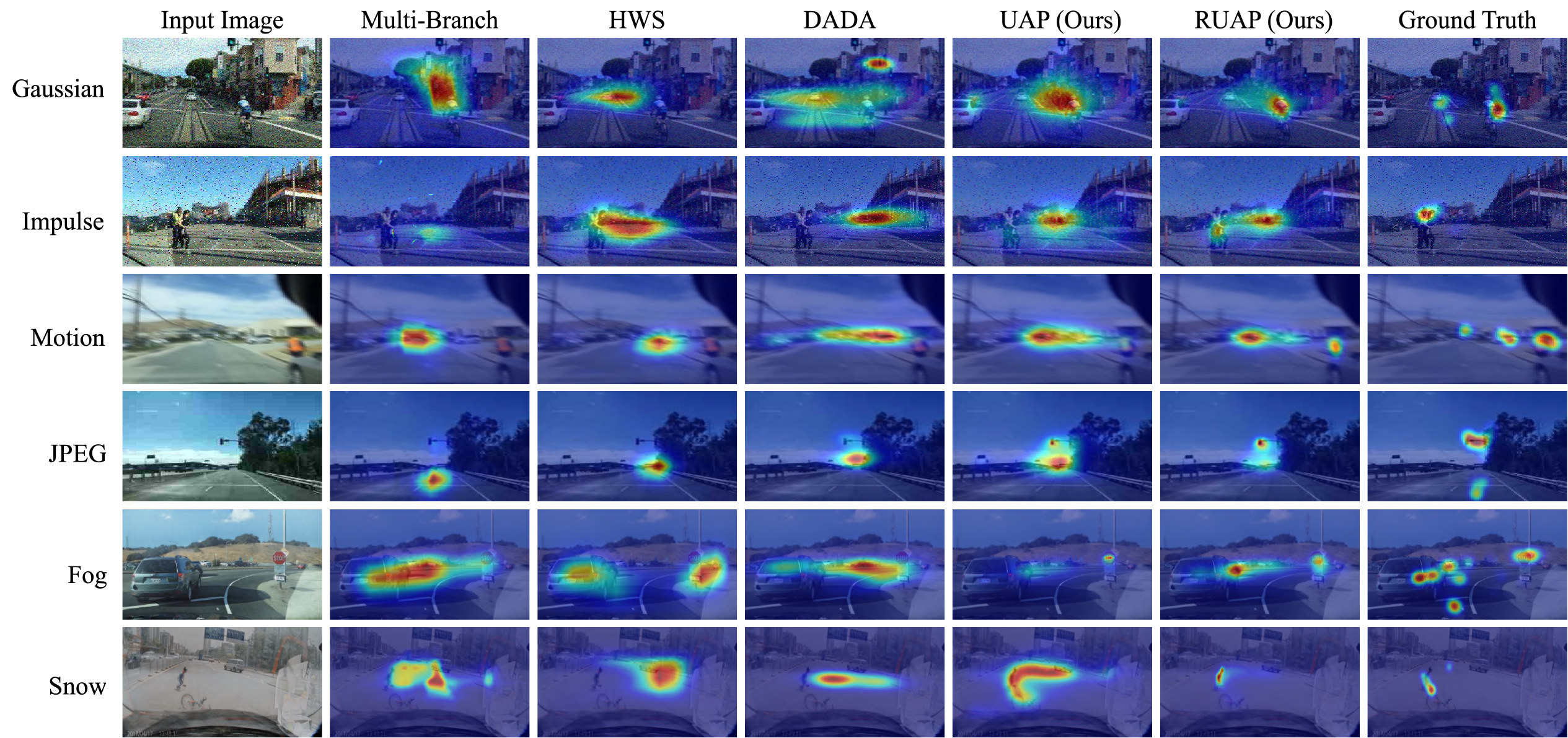}
    \vspace{-2mm}
    \caption{The visualization comparison highlights the performance of our proposed method against other methods under various corruptions. Each row showcases scenarios affected by different types of corruption. It can be observed that our RUAP method effectively focuses on most key regions.}
    \label{fig:cor_vis}
    \vspace{-1mm}
\end{figure*}

\textbf{Corruption Robustness.}~Figure~\ref{fig:cor_vis} shows the visualization comparison of our proposed method against other methods under various corruptions. It can be observed that our UAP method effectively focuses on most key regions. For example, in the first row, under Gaussian noise corruption, only the UAP and Multi-Branch~\cite{palazzi2018predicting} methods successfully capture attention on the bicycles, while other methods fail to estimate the attention correctly. However, both UAP and Multi-Branch exhibit overly dispersed attention on the bicycles. Our further proposed RUAP method resolves this issue by predicting more precise attention regions. In the second row, under impulse noise corruption, only the RUAP method accurately predicts attention on pedestrians. Similar patterns are observed in other scenarios, demonstrating the superiority of our proposed method.

\begin{figure*}
    \centering
    \includegraphics[width=0.92\linewidth]{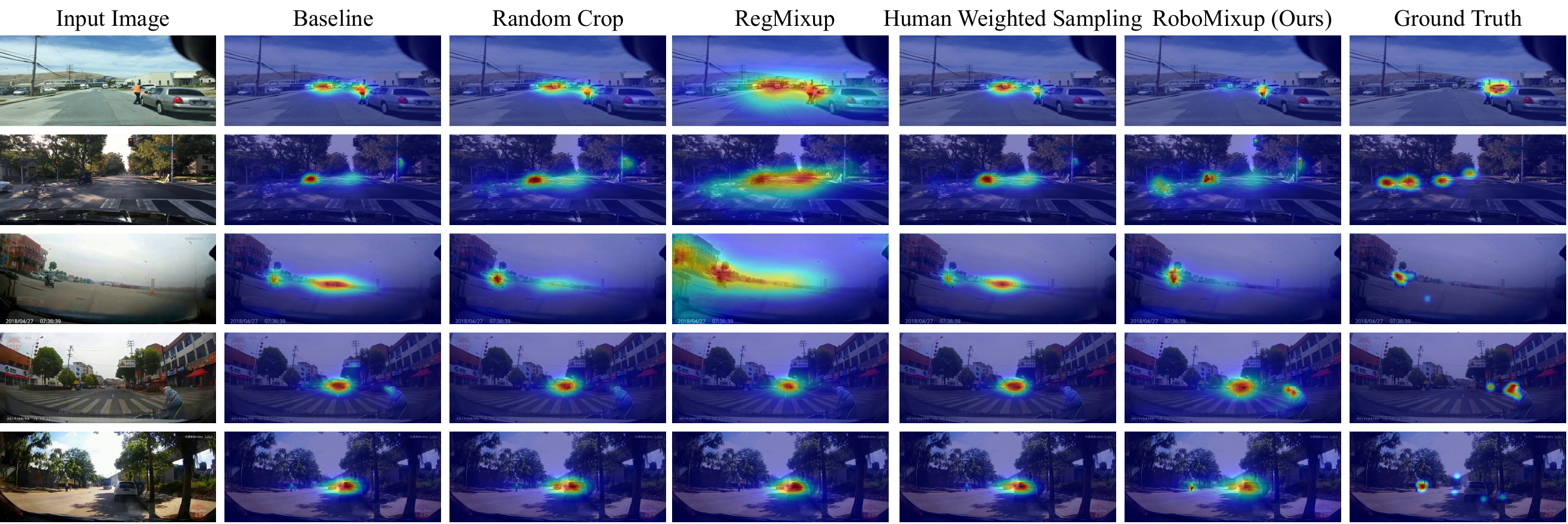}
    \vspace{-2mm}
    \caption{Visualization results obtained by training our proposed unsupervised model using various data augmentation methods. Each row respectively showcases emergency scenarios of autonomous vehicles. The results depict how each model predicts driver attention in these scenarios.}
    \label{fig:crucial_comparison}
    \vspace{-1mm}
\end{figure*}

\textbf{Robustness against central bias.}~We also present a visual comparison of our proposed model, trained with various data augmentation techniques in addressing the central bias. As illustrated in Figure~\ref{fig:crucial_comparison}, the model trained using our proposed RUAP with RoboMixup, is better able to significantly focus on the agents that require attention in critical and challenging scenarios of emergent events compared to other data augmentation methods. This approach more closely aligns with actual driver attention patterns, resulting in improved attention prediction outcomes. To be specific, in the first and third rows, compared to other methods, the RUAP method shows the smallest central bias, correctly focusing attention on the driver alighting from the vehicle and the oncoming motorcycle, respectively. The fourth row displays a scene of a cyclist accident in front of the car, where only our method significantly notices the fallen individual. Failing to correctly place attention on these scenes could lead to severe accidents.




\subsection{Ablation Studies}\label{sec:ablation}
\label{subsec:abl}


\noindent\textbf{Impact of different modules.} Table~\ref{table:ablated} assesses each component of our model to confirm its impact. APB using generated pseudo-labels results from BDD-A results in the lowest performance. Incorporating UMB with its multiple branches significantly enhances model performance, surpassing that of APB alone. Additionally, integrating the non-local block yields a notable improvement. The addition of KEB significantly improves the model, aligning our full model's results with those of leading fully supervised models. 


\noindent{\bf Different source of pseudo-labels.} We evaluated the impact of various pseudo-label sources on outcomes, comparing their performance as shown in Table~\ref{table:number}. The initial rows show training results with single-source pseudo-labels ((\textit{e.g.}, ML-Net or UNISAL). The third row reveals optimal outcomes when combining two pseudo-label sources (ML-Net+UNISAL), highlighting how our UMB improves performance by leveraging multiple pseudo-label interactions. However, utilizing more than two pseudo-label sources leads to decreased performance, as detailed in the following lines. Consequently, we utilize two pseudo-label sources (ML-Net and UNISAL) across our experiments.

\noindent{\bf Domain gap.} Transferring a model directly from one domain to another often results in poor outcomes due to significant domain gaps. Our experiments explicitly demonstrate this phenomenon in Table \ref{table:gap}, which shows a significant performance drop when the APB model is trained on the first two datasets, but tested on SALICON. This suggests a domain gap between self-driving and natural scene attention datasets. Likewise, training the APB on SALICON results in poor performance on self-driving attention datasets. 

\begin{figure}[t]
    \centering
    \includegraphics[width=0.95\linewidth]{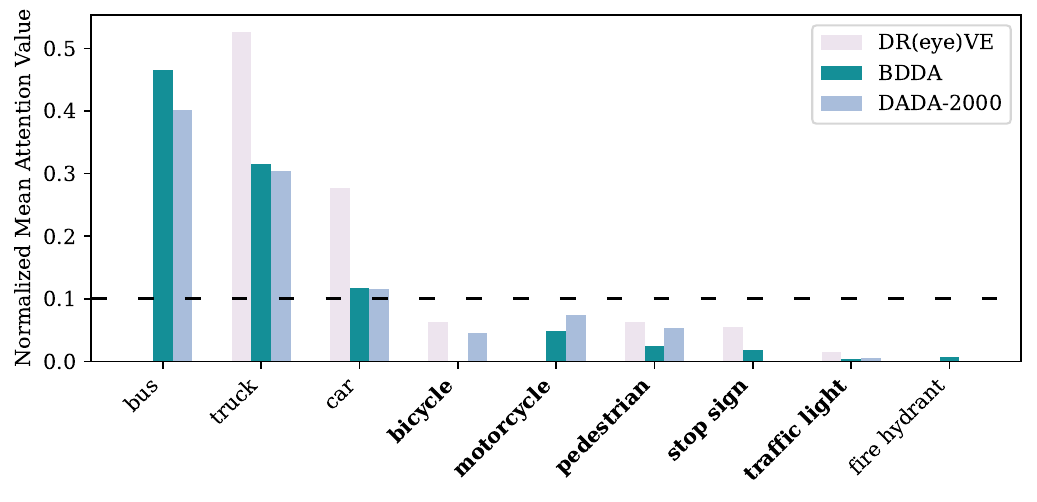}
    \vspace{-5mm}
    \caption{Visualization of the distribution of normalized mean attention values for frequently occurring objects across three datasets. The x-axis represents the frequently occurring objects in the three datasets, while the y-axis shows the normalized attention values for each category. The dashed line indicates the proportion factor \( \eta \), with values below it corresponding to the categories mined as priors. We highlight the manually selected prior categories—bicycles, motorcycles, pedestrians, stop signs, and traffic lights, while excluding 'fire hydrants' as it appears in only one dataset and lacks generalizability.}
    \label{fig:prior_mining}
\end{figure}

\noindent{\bf Prior knowledge embedding.} Our model's Knowledge Embedding Block (KEB) migrates self-driving and traffic knowledge to refine pseudo-labels generated by models pre-trained on natural scenes. However, challenges persist: 1) Which prior traffic knowledge should be incorporated? 2) How should this knowledge be integrated with the generated pseudo-labels? We visualize the normalized mean attention values in Figure~\ref{fig:prior_mining}. Taking the DR(eye)VE~\cite{alletto2016dr} dataset as an example, our mining strategy identifies objects such as bicycles, pedestrians, stop signs, and traffic lights, resulting in distinct attention distributions compared to other objects. Table~\ref{table:way} shows that segmenting key traffic elements alone yields the best results. Additionally, we explore two methods of integrating prior knowledge in KEB: concatenation along the channel dimension ("concat") and the operation defined in Eq.~(\ref{eqo:prior}) ("single"). According to Table~\ref{table:way}, the operation defined in Eq.~(\ref{eqo:prior}) proves to be the most effective.

\noindent{\bf Semi-supervised setting.} Additionally, we compare semi-supervised settings based on the same network as detailed in~\cite{44873}, with outcomes shown in Table~\ref{table:semi}. We implement two semi-supervised approaches: 1) \textbf{Semi-supervised v1} trains the APB on ${1}/{4}$ of BDD-A's randomly sampled labeled data, followed by full network training with pseudo-labels from the remaining images; 2) \textbf{Semi-supervised v2} follows the reverse process. However, Table~\ref{table:semi} reveals significant performance declines in both semi-supervised versions compared to the fully-supervised APB, underperforming even our unsupervised model. The diminished performance is attributed to the use of a small dataset portion, leading to a restricted central bias in the model, particularly within self-driving contexts. Our unsupervised approach benefits from uncertainty mining, transferring generalized information from natural scenes and reducing bias.

\begin{table*}[]
\setlength\tabcolsep{3.6mm}
\centering
\caption{Results comparison of APB trained on different datasets and tested on another dataset. Note that BDD-A, DR(eye)VE, and DADA-2000 are self-driving benchmarks, while SALION is a natural scene dataset. The best results are highlighted in bold.}
\vspace{-2mm}
\label{table:gap}
\begin{tabular}{c|ll|ll|ll|ll}
\toprule
\textbf{Dataset} & \multicolumn{2}{c|}{BDD-A~\cite{xia2018predicting}}      & \multicolumn{2}{c|}{DR(eye)VE~\cite{palazzi2018predicting}}  & \multicolumn{2}{c|}{DADA-2000~\cite{fang2021dada}}  & \multicolumn{2}{c}{SALICON~\cite{jiang2015salicon}}     \\ 
 &
  \multicolumn{1}{c}{KLD↓} &
  \multicolumn{1}{c|}{CC↑} &
  \multicolumn{1}{c}{KLD↓} &
  \multicolumn{1}{c|}{CC↑} &
  \multicolumn{1}{c}{KLD↓} &
  \multicolumn{1}{c|}{CC↑} &
  \multicolumn{1}{c}{KLD↓} &
  \multicolumn{1}{c}{CC↑} \\ 
  \midrule\midrule
BDD-A             & \textbf{1.036} & \textbf{0.657} & \textbf{1.870} & \textbf{0.535} & 1.824          & 0.447          & 1.584          & 0.318          \\
DADA-2000         & 1.357          & 0.543          & 2.044          & 0.484          & \textbf{1.604} & \textbf{0.504} & 1.661          & 0.351          \\
SALICON           & 2.109          & 0.287          & 2.735          & 0.277          & 2.589          & 0.247          & \textbf{0.722} & \textbf{0.552} \\ \bottomrule
\end{tabular}
\vspace{-2mm}
\end{table*}

\begin{table}[!t]
\centering
\caption{Performance comparison of different sources of pseudo-labels in the UMB. We use the following abbreviations: M for ML-Net~\cite{cornia2016deep}, U for UNISAL~\cite{droste2020unified}, T for TASED-Net~\cite{min2019tased}, and S for SAM~\cite{cornia2018predicting}. }
\vspace{-2mm}
\label{table:number}
\setlength\tabcolsep{7.5mm}
\begin{tabular}{l|cc}
\toprule
\textbf{Pseudo-labels}       & \multicolumn{1}{c}{\textbf{KLD↓}} & \multicolumn{1}{c}{\textbf{CC↑}} \\ \midrule
M        & 1.233                    & 0.608                   \\
U          & 1.246                    & 0.597                   \\
M+U & \textbf{1.099}           & \textbf{0.635}          \\
M+U+T & 1.189                    & 0.619                   \\
M+U+S           & 1.162                    & 0.621                   \\
M+U+T+S           & 1.167                    & 0.620                    \\ 
\bottomrule
\end{tabular}
\vspace{-2mm}
\end{table}


\begin{table}
\centering
\caption{Comparison of different strategies and types of knowledge embedding, where ``obj." refers to the masks of objects with Mask-RCNN, and ``text" refers to the masks of text with EAST in the traffic scene.}
\vspace{-2mm}
\label{table:way}
\setlength\tabcolsep{3.9mm}
\begin{tabular}{l|cc}
\toprule
\textbf{Input}                            & \textbf{KLD↓}  & \textbf{CC↑}   \\ 
\midrule
concat (obj. \& text) & 1.126 & 0.626 \\ 
concat (obj.)          & 1.123 & 0.628 \\
single (obj. \& text, $\alpha=0.8$)            & 1.123 & 0.631 \\
single (obj., $\alpha=0.8$)            & \textbf{1.099} & \textbf{0.635} \\
\bottomrule
\end{tabular}
\end{table}

\begin{table}[t]
\centering
\caption{Comparing different training paradigms, \emph{i.e.,} supervised, semi-supervised and unsupervised settings.}
\vspace{-3mm}
\setlength\tabcolsep{5.9mm}
\begin{tabular}{l|cc}
\toprule
\textbf{Training strategy} & \textbf{KLD↓}  & \textbf{CC↑}   \\ \midrule
fully-supervised APB  & \textbf{1.039} & \textbf{0.657} \\
semi-supervised v1   & 1.669 & 0.422 \\
semi-supervised v2   & 1.130 & 0.629 \\
unsupervised      & 1.099 & 0.635 \\ 
\bottomrule
\end{tabular}
\label{table:semi}
\end{table}

\subsection{Discussion}

\emph{Can driver attention prediction enhance the performance and safety of autonomous driving systems?} We explore this question through a simplified autonomous driving decision-making task, which is an interpretable multi-label decision making task proposed by Xu \textit{et al.}~\cite{xu2020explainable}. Specifically, this task involves decision classification across four categories, such as turn-type decisions and corresponding classification reasons. Hence we design a framework that incorporates attention prediction into this decision-making process. In particular, we identify regions of interest based on driver attention and extract features for classification using a graph convolutional network. Experimental results demonstrate that predicting driver attention is crucial to understand the current driving state and environment. By integrating this attention data, the model is better able to perceive and comprehend the driving environment, leading to more interpretable outputs. Further details about the proposed framework and dataset can be found in our supplementary materials.

\section{Conclusion}
This paper presented a novel robust unsupervised approach for predicting attention in self-driving. We showed an uncertainty mining branch and a knowledge embedding block to create reliable pseudo-labels and bridge domain gaps. Furthermore, we released a new dataset \emph{DriverAttention-C} and designed RoboMixup data augmentation to overcome corruption and central bias. Our method's effectiveness, robustness, and superiority are demonstrated through extensive experiments across four benchmarks. Future work will integrate our method into a vision-language large model-based autonomous driving system.

\ifCLASSOPTIONcaptionsoff
  \newpage
\fi



%
{\small
\bibliographystyle{unsrt}
\bibliography{Main.bbl}
}

%








\end{document}